\RequirePackage{tikz}
\tikzset{
  treenode/.style = {shape=rectangle, rounded corners,
                     draw, align=center,
                     top color=white, bottom color=blue!20},
  root/.style     = {treenode, font=\Large, bottom color=red!30},
  env/.style      = {treenode, font=\ttfamily\normalsize},
  dummy/.style    = {circle,draw, align=center, minimum width=0.8cm}
}
\usetikzlibrary {arrows.meta}
\documentclass[pdflatex,sn-mathphys]{sn-jnl}

\usepackage{lineno, hyperref}
\usepackage{graphicx}
\usepackage[misc]{ifsym}

\usepackage{amsmath}
\usepackage{bm}
\usepackage{amssymb}
\usepackage{caption}
\usepackage{paralist}

\usepackage{algpseudocode}
\usepackage{verbatim}

\usepackage{float}

\usepackage[utf8]{inputenc} 
\usepackage[T1]{fontenc}    
\usepackage{booktabs}       
\usepackage{amsfonts}       
\usepackage{nicefrac}       
\usepackage{microtype}      
\usepackage{url}
\usepackage{moresize}

\usepackage[labelformat=simple]{subcaption}

\usepackage{amsthm}

\newtheorem{remark}{Remark}

\jyear{2022}%

\begin{document}

\title[Loss-Optimal Classification Trees]{Loss-Optimal Classification Trees: A Generalized Framework and the Logistic Case}


\author*[1]{\fnm{Tommaso} \sur{Aldinucci}}\email{tommaso.aldinucci@unifi.it}

\author[1]{\fnm{Matteo} \sur{Lapucci}}\email{matteo.lapucci@unifi.it}

\affil[1]{\orgdiv{Dipartimento di Ingegneria dell'Informazione}, \orgname{University of Florence}, \orgaddress{\street{Via di Santa Marta 3}, \city{Florence}, \postcode{50139}, \country{Italy}}}


\abstract{Classification Trees are one of the most common models in interpretable machine learning. Although such models are usually built with greedy strategies, in recent years, thanks to remarkable advances in Mixer-Integer Programming (MIP) solvers, several exact formulations of the learning problem have been developed. In this paper, we argue that some of the most relevant ones among these training models can be encapsulated within a general framework, whose instances are shaped by the specification of loss functions and regularizers. Next, we introduce a novel realization of this framework: specifically, we consider the logistic loss, handled in the MIP setting by a linear piece-wise approximation, and couple it with $\ell_1$-regularization terms. The resulting Optimal Logistic Classification Tree model numerically proves to be able to induce trees with enhanced interpretability properties and competitive generalization capabilities, compared to the state-of-the-art MIP-based approaches.
}




\maketitle

\section{Introduction}\label{sec1}

\textit{Classification Trees} (CTs) are a popular machine learning model for classification problems  \cite{kotsiantis2013decision,song2015decision}. Introduced in a seminal work by Breiman et al.\ in 1984 \cite{breiman84classification}, CTs have widely been employed for decades, especially for tasks with small-sized, tabular data.

Formally, a CT is a connected acyclic graph whose nodes are linked in a hierarchical manner. Each internal node (\textit{branch}) acts by splitting data in the feature space,  based on some predefined condition. Finally, nodes that have no children (i.e., there is no outbound arc connecting them to lower nodes in the hierarchy) are called \textit{leaves} and are associated with a prediction label for samples that are forwarded to them by the above branches.

Most algorithms designed to construct classification trees define \textit{univariate} (axis-aligned) \textit{splits} \cite{breiman84classification,quinlan1986induction,de1991distance}. With univariate splits, a single feature is chosen at each node and the corresponding value of each data point is compared to a given threshold; data points are thus forwarded to one of the children nodes, based on the result of this comparison. 
Each prediction of the model is therefore finally obtained following a path along the tree, based on the sequence of ``decisions'' at each encountered node. 

In order to improve the expressive power of CTs, however, more complex splitting rules have been also considered in the literature for defining branching splits. These rules can take into account linear (in this case we talk about \textit{oblique} trees) or even nonlinear relations between features \cite{friedman1977recursive, loh1988tree, john1995robust}. 

Yet, the intrinsic interpretability of CTs is a key factor that makes them particularly popular in domains, such as healthcare, finance or justice, where understanding the decision-making process and accounting for the choices made is important \cite{Rudin2019Stop}. 
For this reason, recent research has dealt with training algorithms to construct CTs having both good predictive performance and intrinsic interpretability properties. In particular, oblique trees are often considered interpretable as long as branching linear classifiers are actually sparse models \cite{ross2017neural, ribeiro2016should, jovanovic2016building}. Then, the rules that ultimately define the decision process are basically ``if-then-else'' conditions with very simple clauses; hence, domain experts with possibly no technical background are still typically able to interpret and understand the logic that leads to decisions. 

Within this scenario, the contribution of the present paper consists of the encapsulation of the concept of \textit{classification loss} within each splitting rule in the popular optimal CTs (OCTs) framework \cite{bertsimas2017optimal}. This point of view also generalizes the idea of \cite{d2022margin}, where \textit{maximum margin} splits were shown to improve
the generalization performance of optimal trees. In fact, both OCTs and margin-optimal trees can be seen as particular instances of the generalized \textit{loss-optimal classification tree} framework discussed in this manuscript.  

Following further this path, we propose to choose the \textit{log loss} to construct \textit{optimal logistic classification trees} (OLCTs). This way, the nice properties of logistic regression models (e.g., generalization, calibration, interpretability) can be combined with the easily readable structure of CTs. We show that the logistic loss can be effectively encapsulated within the standard \textit{mixed-integer linear programming} (MILP) models for OCTs, using the piece-wise linear approximation defined in \cite{sato2016feature}, and thus handled by the usual off-the-shelf MILP solvers.

In addition, we point out a simple patch for the soft feature selection strategy employed in \cite{d2022margin} for maximum margin CT models: the sparsity requirement can be more easily handled exploiting $\ell_1$-regularization; clearly, this idea can then be extended to the case of general loss-optimal models and thus to the OCLTs framework, which finally allows to obtain sparse, yet effective classification models with the additional interpretability properties of logistic regression.

The rest of the manuscript is organized as follows: in Section \ref{sec:literature} we report a detailed review of the literature concerning classification trees induction. Then, in Section \ref{sec:octs}, we derive the general MIP framework for optimal classification trees, based on the concepts of loss and regularizer. We consequently introduce in Section \ref{sec:olcts} a novel specification of the general framework, leading to the definition of the Optimal Logistic Classification Tree model. In Section \ref{sec:exp}, we then report the results of computational experiments aimed at assessing the actual potential of the proposed approach. Finally, we give some concluding remarks in Section \ref{sec:conclusion}.

\section{Related literature}
\label{sec:literature}
The construction of an optimal decision tree is an $\mathcal{NP}$-complete problem \cite{laurent1976constructing} and, for this reason, traditional algorithms apply top-down greedy strategies \cite{rokach2005top}, employing quality measures to define the best parameters for each branch node \cite{breiman84classification,quinlan1986induction,de1991distance}.

Easily applicable even with large datasets, unfortunately greedy approaches often generate sub-optimal structures with poor generalization capabilities \cite{murthy1995decision, norouzi2015efficient}. This is mainly due to the fact that decision rules at the deepest nodes are often defined taking into account very small portions of the dataset. Once the tree has been grown, these methods usually apply a post-pruning phase to reduce the complexity of the structure and alleviate the overfitting problem \cite{breiman84classification, quinlan1987simplifying, bohanec1994trading}.

In order to overcome these performance drawbacks, several methods have been proposed in the literature. One of the most common approaches consists of resorting to ensembles to reduce the model variance. Particularly relevant techniques of this kind are Random Forest \cite{breiman2001}, Gradient Boosting Machines \cite{friedman2001greedy} and XGBoost  \cite{chen2016xgboost} which combine, in different ways, multiple trees to increase the overall performance. 

Other works are focused on \textit{oblique} classification trees, i.e., on the usage of hyperplanes to recursively divide the feature space. In this case, the tree model exploits a \textit{multivariate} linear function at each branch node so that multiple features are involved in the decision process \cite{murthy1994system,brodley1995multivariate,orsenigo2003multivariate}.
Although both ensembles and oblique trees are more expressive than standard univariate classification trees, both classes of approaches suffer from an evident loss of interpretability.

A conceptually different path to improve the performance concerns the improvement of the fitting procedure, rather than the development of more expressive models. In particular, exact formulations of the learning problem exploiting linear programming have been considered. Back in the early '90s, Bennet et al. proposed linear methods for constructing separation hyperplanes \cite{bennett1992robust,bennett1994multicategory} and for the induction of oblique classification trees solving an LP problem at each branch node \cite{bennett1992decision}.

More recently, thanks to the outstanding improvements in both hardware power and software solvers capabilities \cite{bixby2012brief}, new formulations of the learning problem have been developed. These formulations are related to both univariate and multivariate splits, and exploit various mathematical optimization techniques \cite{carrizosa2021mathematical} for an effective modelling of the learning problem. One particularly prominent approach involves the use of Mixed Integer Linear Programming (MILP) formulations. The seminal work in this context is the one by Bertsimas and Dunn \cite{bertsimas2017optimal}, where an exact MILP model is proposed that can be solved up to a certifiably globally optimal CT (OCT) structure (in terms of misclassification error). Inspired by this work, a plethora of improved MILP-based approaches followed; among them, we can notably cite a formulation of the problem for the case of binary classification with categorical features \cite{gunluk2021optimal}, or a flow-based formulation exploiting linear relaxation and Bender's decomposition to efficiently handle the specific case of binary classification with binary features \cite{aghaei2021strong}. A reformulation for the OCT model was also proposed for the case of parallel splits, allowing to significantly improve the efficiency  of the approach \cite{verwer2019learning}.
Focusing on the interpretability aspect of CTs, some works take advantage of regularization terms to limit the number of branch nodes \cite{bertsimas2017optimal} and total number of leaves in the model \cite{hu2019optimal,lin2020generalized}.

The above formulations can in principle generate certified globally optimal structures. In practice, however,  there is a combinatorial explosion in the number of binary variables as the depth of the tree or the size of the dataset increases. Thus, optimality gap can be closed by branch-and-bound type procedures only for really shallow trees and on tasks with a few hundred examples at most.

For this reason, local optimization strategies have also been investigated. Starting from a given tree, trained by a greedy approach, these methods refine its structure by iteratively minimizing the misclassification loss associated with each node in the tree. Carreira-Perpin{\'a}n and Tavallali \cite{carreira2018alternating} introduce an alternating minimization strategy, that operates level-wise on the tree, decomposing the learning problem and solving each sub-problem up to global optimality. Similarly, Dunn proposed Local Search methods \cite{dunn2018optimal}, able to handle real world datasets and to refine greedy structures grown with CART-like algorithms, at the cost of producing suboptimal models. 

In addition to exploring integer optimization, researchers have also delved into continuous optimization approaches within the context of optimal trees. Blanquero et al. presented a nonlinear programming model in their work \cite{blanquero2021optimal}, aiming to develop an optimal "randomized" classification tree with oblique splits. This approach involves making random decisions at each node based on a soft rule, which is induced by a continuous cumulative density function.

Expanding on their earlier findings, Blanquero et al. addressed the issues of global and local sparsity in the randomized optimal tree model in \cite{blanquero2020sparsity}. To tackle these challenges, they incorporated regularization terms based on polyhedral norms. By employing this regularization technique, the researchers aimed to promote sparse solutions both globally and locally within the randomized tree framework.
The same approach has also been investigated for the regression case in \cite{blanquero2022sparse}.

It is important to note that, in the randomized framework proposed by Blanquero et al., the assignment of a sample to a specific class is not deterministic. Instead, it is determined based on a given probability, allowing for a more flexible and probabilistic classification approach.

A different stream of research aims at improving the performance of classification trees exploiting the concept of loss in defining splits. Specifically, there are works that deal with oblique splits, greedily computing at each branch node the parameters of the separating hyperplane as a Support Vector Machine (SVM) \cite{cortes1995support}.
The formulation from \cite{bennett1998support} for SVM branches exploits a dual convex quadratic model that can also handle kernel functions to capture non linear patterns in data.
Tibshirani and Hastie instead developed a greedy algorithm to handle high dimensional features and multiclass classification, inducing classification trees with margin properties, i.e., structures where each branch node employs a linear SVM to split the feature space \cite{tibshirani2007margin}.

With reference again to multivariate trees, approaches that exploit logistic models at nodes have also been studied. In \cite{landwehr2005logistic} a method is presented that induces standard axis-aligned classification trees with the difference that, at each leaf node, a logistic regression model is built to provide the final prediction. Moreover, a greedy method to fit a piecewise linear logistic regression model is presented in \cite{chan2004lotus}, exploiting decision trees to recursively partition the feature space. The final model can be basically seen as a multivariate classification tree with a logistic regressor on each branch node.

Recently, D'Onofrio et al. \cite{d2022margin} took advantage of an exact mixed-integer quadratic programming (MIQP) formulation of the CT learning problem including the concept of maximum margin. This approach allows to define OCTs where each branch node is a maximum-margin linear classifier.
The maximum margin property, obtained minimizing the hinge loss at all branch nodes, allows for significant improvements on generalization performance. However, relying on oblique splits has the usual drawback of undermining interpretability. For this reason, the authors also proposed strategies (both hard and soft) to induce sparsity in the coefficients of each hyperplane. 


\section{Generalized Loss-Optimal Classification Trees}
\label{sec:octs}
\subsection{Notation and Oblique Tree Structure by Mixed-Integer Programming}
Let $\mathcal{I} = \{1,..., n\}$ and $\mathcal{D} : = \{ (\boldsymbol{x}_i, y_i), \ \boldsymbol{x}_i \in \mathbb{R}^p , \ y_i \in \{-1, 1 \} , \ i \in\mathcal{I} \}$ be a finite dataset of $n$ observations with $p$ features and binary labels.

We introduce the notation, largely based on \cite{bertsimas2017optimal, d2022margin}, used for the formulation of the Loss-Optimal CTs learning problem.
Formally, let $\mathcal{T}$ be the set of branch nodes of a generic (oblique) CT. We denote by $d$ the number of layers of branching nodes of the tree, i.e., its depth, by $\mathcal{H}$ the set $\{1,\ldots,d\}$ and by $\mathcal{T}_{h}$ the set of nodes at depth $h$, for $h\in\mathcal{H}$.

Each branch node $t \in \mathcal{T}$ is characterized by a linear function $\boldsymbol{w}_t^T \boldsymbol{x} + b_t, \ \boldsymbol{w}_t \in \mathbb{R}^p, \ b_t \in \mathbb{R} $, which induces the hyperplane $\boldsymbol{w}_t^T \boldsymbol{x} + b_t = 0$ as decision boundary. This has the effect of forwarding an examined data point $\boldsymbol{x}_i$ to the left child of node $t$ if $\boldsymbol{w}_t^T\boldsymbol{x}_i + b_t \leq 0$ and to the right child otherwise.
Note that, using this notation, a glass-box axis-aligned CT can be obtained as a special case, imposing  $\lVert\boldsymbol{w}_t\rVert_0 = 1, \ \forall \ t \in \mathcal{T}$, where $\|\cdot\|_0$ denotes the $\ell_0$ pseudo-norm.

\begin{figure}

\begin{center}
    \begin{tikzpicture}[ level distance=1.5cm,
      level 1/.style={sibling distance=6cm},
      level 2/.style={sibling distance=3cm},
      level 3/.style={sibling distance=2.0cm}]
    \node[circle, draw, minimum width=0.60cm, label={182:$\boldsymbol{w}_0^T\boldsymbol{x}+b_0 \leq 0$}, label={358:$\boldsymbol{w}_0^T\boldsymbol{x}+b_0 > 0$}](0){\scalebox{.6}{0}}
      child{
        node[circle,draw, minimum width=0.60cm, label={183:$\boldsymbol{w}_1^T\boldsymbol{x}+b_1 \leq 0$}, label={357:$\boldsymbol{w}_1^T\boldsymbol{x}+b_1 > 0$}, fill=white, text = black ](1){\scalebox{.6}{1}}
        child{
            node[circle,draw, minimum width=0.60cm, fill=white, text=black, label={183: $\leq $}, label={357:$> $}](3) {\scalebox{.6}{3}}
            child{node[circle,draw, minimum width=0.60cm, fill=gray, text = black, label={268: $-1 $}] {\scalebox{.6}{7}}}
            child{node[circle,draw, minimum width=0.60cm, fill=gray, text = black, label={268: $1 $}] {\scalebox{.6}{8}}}} 
        child{
            node[circle,draw, minimum width=0.60cm, fill=white, text=black, label={183: $\leq $}, label={357:$> $}](4) {\scalebox{.6}{4}}
            child{node[circle,draw, minimum width=0.60cm, fill=gray, text = black, label={268: $-1 $}] {\scalebox{.6}{9}}}
            child{node[circle,draw, minimum width=0.60cm, fill=gray, text = black, label={268: $1 $}] {\scalebox{.6}{10}}}} 
        } 
      child{
        node[circle,draw, minimum width=0.60cm, label={183:$\boldsymbol{w}_2^T\boldsymbol{x}+b_2 \leq 0$}, label={357:$\boldsymbol{w}_2^T\boldsymbol{x}+b_2 > 0$}]{\scalebox{.6}{2}} 
        child{
            node[circle,draw, minimum width=0.60cm, fill = white, text = black, label={183: $\leq $}, label={357:$> $}](5) {\scalebox{.6}{5}}
            child{node[circle,draw, minimum width=0.60cm, fill=gray, text = black, label={268: $-1 $}] {\scalebox{.6}{11}}}
            child{node[circle,draw, minimum width=0.60cm, fill=gray, text = black, label={268: $1 $}] {\scalebox{.6}{12}}} 
        }
        child{
            node[circle,draw, minimum width=0.60cm, fill=white, label={183: $\leq $}, label={357:$> $}](6) {\scalebox{.6}{6}}
            child{node[circle,draw, minimum width=0.60cm, fill=gray, text=black, label={268: $-1 $}] {\scalebox{.6}{13}}}
            child{node[circle,draw, minimum width=0.60cm, fill=gray, text=black, label={268: $1 $}] {\scalebox{.6}{14}}}
        }
        };
    \end{tikzpicture}
\end{center}

\vspace{1em}
\caption{An oblique tree of depth 3. Branches are on white background, leaves on gray. }
\label{fig:Tree1}
\end{figure}
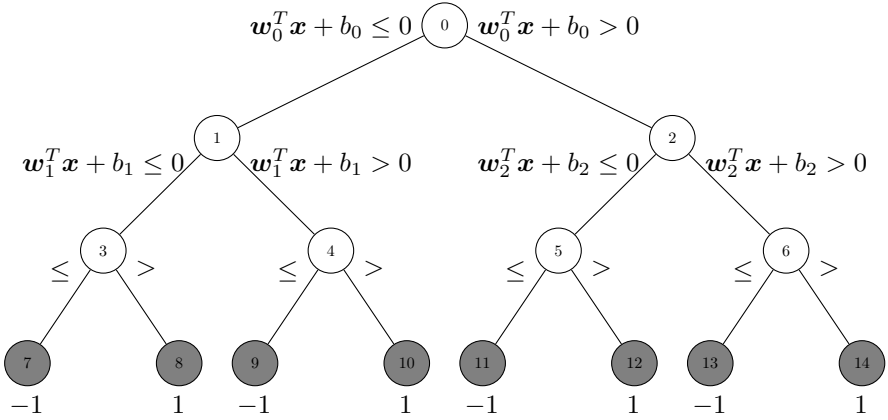

Since we are interested in CTs where each node is a binary classifier, we can let the final prediction of the point depend solely on the decision of the last branch node. Indeed, each splitting hyperplane is a linear classifier with decision function $\text{sgn}(\boldsymbol{w}_t^T\boldsymbol{x} + b)$.
In view of this, there is no need of tracking the points up until the \textit{leaves}, so that the number of nodes to be modelled can be reduced of $2^d$ units, or, in other words, cut to a half. An example of an oblique CT with branching classifiers of depth 3 is shown in Figure \ref{fig:Tree1}.



We now introduce the essential building blocks of the MILP models proposed in the literature for OCTs training.  
Recalling that $\mathcal{T}_{d} \subset \mathcal{T}$ is the set of nodes of the last branching layer, the routing of each point to the proper branch of the last layer can be modelled introducing the binary variables:
\begin{equation*}
    z_{i, t} = 
    \begin{cases}
         1 & \text{if example} \ \boldsymbol{x}_i \ \text{reaches node} \ t \in \mathcal{T}_d,\\
         0 & \text{otherwise}.
    \end{cases}
\end{equation*}

Using these variables, we can force each point to be routed to exactly one of the last branching nodes, complying with the structure of the model. In particular, this can be obtained imposing the following constraints:
\begin{subequations}
\label{z}
    \begin{align}
    \sum_{t \in \mathcal{T}_d} z_{i,t} = 1  &\qquad\qquad \forall \ i \in \mathcal{I}, \label{sumz}\\
    z_{i,t}\in\{0,1\}&\qquad\qquad\forall i\in \mathcal{I}, \;t\in\mathcal{T}_d. \label{zbin}
\end{align}    
\end{subequations}

We then introduce the constraints to take into account the path of each data point, which is fully determined by the sequence of decisions taken at the branch nodes.
For this purpose, we introduce the set $\mathcal{T}_d(t) \subseteq \mathcal{T}_d$ as the set of branches at the last layer that are successors of the node $t$; in other words, nodes in $\mathcal{T}_d(t)$ belong to the last layer of the sub-tree rooted at $t$; moreover, we define the subsets $\mathcal{T}_d^r(t)$, $\mathcal{T}_d^l(t)$ as a partition of $\mathcal{T}_d(t)$ such that:
\begin{itemize}
    \item $\mathcal{T}_d^r(t)$ is the set of nodes of the last branching level that belong to the right sub-tree rooted at node $t$;
    \item $\mathcal{T}_d^l(t)$ is the set of nodes of the last branching level that belong to the left sub-tree rooted at node $t$.
\end{itemize}

Each branch $t$ forwards points to its left child if $\boldsymbol{w}_t^T\boldsymbol{x}+b_t \leq 0$ and to the right one otherwise.
This logic can be enforced by means of the following forwarding constraints:
\begin{subequations}
    \label{Forward}
    \begin{align}
    \boldsymbol{w}_t^T\boldsymbol{x}_i+b_t \leq M\Bigl (1 - \sum_{s \in \mathcal{T}_d^l(t)} z_{i, s} \Bigr)  &\qquad\qquad \forall \ i \in \mathcal{I}, \ \forall \ t \in \mathcal{T} \setminus \mathcal{T}_d, \label{leftForward}\\
    \boldsymbol{w}_t^T\boldsymbol{x}_i+b_t \geq \epsilon - M\Bigl(1 - \sum_{s \in \mathcal{T}_d^r(t)} z_{i, s}\Bigr)  &\qquad\qquad \forall \ i \in \mathcal{I}, \ \forall \ t \in \mathcal{T} \setminus \mathcal{T}_d, \label{rightForward}
\end{align}
\end{subequations}
where $M$ is a suitable constant for a big-$M$ type constraint and $\epsilon$ is a small constant preventing numerically degenerate cases.

For example, given $t \in \mathcal{T}$ and a data point $\boldsymbol{x}_i$, if $  z_{i,s} = 1$ and $s \in \mathcal{T}_d(t)$, then $\boldsymbol{x}_i$ has to be routed to the last branch node $s$, which either belongs to $\mathcal{T}_d^l(t)$ or $\mathcal{T}_d^r(t)$. Hence, one and only one of the following pair of equations hold:
$$\sum_{s \in \mathcal{T}_d^l(t)} z_{i, s} = 1,\qquad \sum_{s \in \mathcal{T}_d^r(t)} z_{i, s} = 1.$$
If the former holds, then \eqref{leftForward} guarantees that $\boldsymbol{w}_t^T\boldsymbol{x}_i+b_t \leq 0$, otherwise \eqref{rightForward} forces $\boldsymbol{w}_t^T\boldsymbol{x}_i+b_t > 0$. The big-$M$ strategy makes the constraint for the ``wrong case'' irrelevant. Also, no constraint applies for branch $t$ in case $\boldsymbol{x}_i$ is forwarded to a node $s \in \mathcal{T}_d \setminus \mathcal{T}_d(t)$.
Moreover, we shall observe that these constraints are not required if $t \in \mathcal{T}_d$ since the last branching level is only responsible to make the final prediction. 

\subsection{The Role of Loss Functions}
For a CT to be meaningful, a suitable \textit{loss function} necessarily has to be used to push the last branching layers and the overall model to correctly classify the data points. This can be done by estimating the errors, or slacks $\xi_i,\,i=1,\ldots,n$, committed on each data point:
\begin{equation}
    \label{eq:loss}
    L=\sum_{i\in I}^{}\xi_i,\qquad \boldsymbol{\xi}\in\mathbb{R}^n.
\end{equation}
The definition of quantities $\boldsymbol{\xi}$ clearly depends on the particular loss function to be used. For example, most MIP models for OCTs  \cite{bertsimas2017optimal,aghaei2021strong,hu2019optimal,carreira2018alternating} directly employ the \textit{misclassification loss}, which is defined using additional binary variables and big-$M$ constraints as

\begin{subequations}
\label{misclass}
    \begin{align}
    \label{misclass_1}
    \boldsymbol{w}_t^T\boldsymbol{x}_i+b_t\le M(2+\hat{y}_i-z_{i,t})&\qquad \forall i\in\mathcal{I},\;\forall\,t\in\mathcal{T}_d,\\ \boldsymbol{w}_t^T\boldsymbol{x}_i+b_t\ge -M(2-\hat{y}_i-z_{i,t})&\qquad\forall i\in\mathcal{I},\;\forall\,t\in\mathcal{T}_d, \label{misclass_2}\\ \label{misclass_3}
     \xi_i = \frac{1}{2}(1-y_i\hat{y}_i)&\qquad\forall i\in\mathcal{I},\\\label{misclass_4}\hat{\boldsymbol{y}}\in\{-1,1\}^n.&
\end{align}
\end{subequations}
The additional variables $\hat{y}$ model the predicted class for each data point; the slack variable corresponding to each data point will be set to 0 if prediction is correct $y_i\hat{y}_i=1$, otherwise it will be equal to 1. Constraints \eqref{misclass_1}-\eqref{misclass_2} guarantee that, for a point $x_i$ arriving at the leaf $t$, $\hat{y}_i=1$ if and only if $\boldsymbol{w}_t^T\boldsymbol{x}_i+b_t\ge 0$. 

However, this is not necessarily the only option for a loss function. In fact, different choices not only might be statistically more robust, but may also avoid the introduction of the additional binary variables and logical constraints that increase problem complexity. An example of a continuous loss with these features, easily embeddable in the MIP framework, is the \textit{hinge loss}, $\max\{0,1-y_i(\boldsymbol{w}^T\boldsymbol{x}_i+b)\}$, that can be modeled as:
\begin{subequations}
\label{slacks_last_layer}
    \begin{align}
    \xi_{i} \geq 1 - y_i(\boldsymbol{w}_t^T\boldsymbol{x}_i+b_t) - M\Bigl(1 -  z_{i, t} \Bigr)  &\qquad\ \forall \ i \in \mathcal{I}, \ \forall \ t \in \mathcal{T}_d, \label{slacks_last_layer_1}\\
    \xi_{i} \geq 0 & \qquad\ \forall \ i \in \mathcal{I}, \ \forall \ t \in \mathcal{T}_d. \label{slacks_last_layer_2}
\end{align}
\end{subequations}

The objective function can then be defined as the sum of the total loss function $L$ and a regularization term for weights $\boldsymbol{w}_t$, $t\in\mathcal{T}_d$, so that we basically have an empirical risk minimization problem: 
\begin{equation}
    \label{eq:obj_last_layer}
     L + \lambda \sum_{t\in\mathcal{T}_d}\Omega(\boldsymbol{w}_t).
\end{equation}
We note that:
\begin{itemize}
    \item  By setting $\lambda=0$ and using \eqref{misclass},  we actually retrieve the standard OCTs problem from \cite{bertsimas2017optimal}; 
    \item setting $\lambda>0$, $\Omega(\cdot) = \|\cdot\|_2^2$ and using \eqref{slacks_last_layer} we get ``SVM leaves''.
\end{itemize}
 
In fact, loss terms can also be associated with the upper branching nodes; in this case, we have $d$ vectors of slack variables, $\boldsymbol{\xi}_1,\ldots,\boldsymbol{\xi}_d$: along its path to the leaves, each data point encounters only one node at each layer, and thus only the slack associated with the corresponding classifier has to be taken into account. The overall objective function for a general \textit{loss-optimal} CT model is therefore given by
\begin{equation}
    \label{gen_obj_full}
    \sum_{h\in \mathcal{H}}\left(\sum_{i\in \mathcal{I}}\xi_{i,h} + \lambda_h\sum_{t\in\mathcal{T}_h}\Omega(\boldsymbol{w}_t)\right).
\end{equation}

It is worth noting at this point that the Margin Optimal CT models (MARGOT) from \cite{d2022margin} can then be retrieved as a special case of the general framework, setting \begin{subequations}
\label{svm_all_layers}
    \begin{align}
    \xi_{i,h}\ge 1-y_i(\boldsymbol{w}_t^T\boldsymbol{x}_i+b_t)- M\Bigl(1 - \!\!\!\sum_{s \in \mathcal{T}_d(t)} \!\!\!z_{i, s} \Bigr)  &\quad \forall \, i \in \mathcal{I},\, \forall\, h\in \mathcal{H},\,  \forall\, t \in \mathcal{T}_h, \label{softMargin}\\
    \xi_{i,h} \geq 0 & \quad \forall \, i \in \mathcal{I}, \, \forall\, h\in\mathcal{H} .\label{slacks}
\end{align}    
and \end{subequations} $\Omega(\cdot)=\|\cdot\|_2^2$.




\subsection{Exact Modelling of $\ell_0$ Terms}
In order to enhance the interpretability of the models, sparsity can be compelled within the weights of branching classifiers, thus inducing features selection. The $\ell_0$ norm of vectors $\boldsymbol{w}_t$ can easily be modeled, in a MINLP program, by introducing binary indicator variables, big-$M$ constraints and linear expressions: 
    \begin{gather}
    \label{l0_bin}
        \boldsymbol{\delta}_t\in\{0,1\}^p, \qquad -M\boldsymbol{\delta}_t \le\boldsymbol{w}_t\le M\boldsymbol{\delta}_t,\qquad \|\boldsymbol{w}_t\|_0 = \boldsymbol{1}^T\boldsymbol{\delta}_t.
    \end{gather}
Then, the value of $\|\boldsymbol{w}_t\|_0$ can either be upper bounded or penalized. Following the terminology in \cite{d2022margin} we talk about Hard Feature Selection (HFS) in the former case and Soft Feature Selection (SFS) in the latter one.

\section{Optimal Logistic Classification Trees}
\label{sec:olcts}
In this section, we formalize a novel, particular instance of loss-optimal CT model, the Optimal Logistic CT (OLCT). In particular, we first show how to introduce the logistic loss function within the considered MIP; then, we point out the benefits of using $\ell_1$-regularization terms; we then show the resulting overall optimization model. 

\subsection{Logistic Loss in Mixed Integer Linear Optimization}
The logistic loss function for binary linear classifiers is defined as
$$\ell(\boldsymbol{w}^T\boldsymbol{x}_i+b;y_i) = f(y_i(\boldsymbol{w}^T\boldsymbol{x}_i+b)) =\log(1+\exp(-y_i(\boldsymbol{w}^T\boldsymbol{x}_i+b))).$$
This loss function appears in the individual terms of the summation in the negative log likelihood function of logistic regression models (see, e.g, \cite{hastie2009elements}), i.e., the linear model for binary classification that is obtained by maximum likelihood estimation under the assumption that data follows a Bernoulli distribution. 

In addition to often having strong performance in terms of out-of-sample prediction accuracy, other advantages of logistic regression compared to other linear classifiers, such as SVMs, include
\begin{enumerate}[(a)]
    \item the possibility of obtaining estimates of features importance by simple manipulation of the model weights \cite{hastie2009elements};
    \item the opportunity of getting calibrated probability estimates associated with predictions.
\end{enumerate} 
The above properties offer nice insights that are valuable from the perspective of model interpretability. These considerations motivate us to consider the employment of the logistic loss within the general framework for loss-optimal CTs, discussed in Section \ref{sec:octs}.

The straightforward objection, at this point, concerns the nonlinearity of the logistic loss function; indeed, contrarily to the hinge loss defining SVMs, the log loss cannot exactly be modeled by linear constraints in MILP. However, this issue can be addressed following the strategy, proposed in \cite{sato2016feature}, where the best subset selection problem in logistic regression is solved by means of a MILP approach. In particular, Sato et al.\ proposed to approximate the logistic loss function by a piece-wise linear underestimator. The function
$$f(v) = \log({1+\exp(-v)})$$
is a convex function, thus its tangent line at a point $v_0$ constitutes a global underestimator; more explicitly, for all $v,v_0\in\mathbb{R}$, it holds
$$f(v)\ge f(v_0)+f'(v_0)(v-v_0).$$

To obtain an accurate approximation of the logistic loss, we can thus construct a piece-wise linear underestimator obtained as the point-wise maximum of a family of tangent lines:
$$f(v)\approx \max\{f(v_k)+f'(v_k)(v-v_k)\mid k=0,\ldots,K\}.$$

Sato et al.\ also propose a greedy strategy to select points $v_k$ where computing the tangent lines so as to minimize the approximation error: at each iteration, a tangent line is added to the piece-wise linear approximation so that the area between the exact and approximated loss function is minimized. The resulting sets of tangent points to be used for increasingly accurate approximations of the logistic loss are
\begin{gather*}
    V_0 =\{0,\pm\infty\},\qquad V_1 = V_0\cup \{\pm1.9\},\qquad V_2=V_1\cup \{\pm0.89,\pm3.55\},\\
    V_3 = V_2\cup\{\pm0.44,\pm1.37,\pm2.63,\pm5.16\}.
\end{gather*}
Obviously, a larger number of points leads to a larger number of constraints in the MILP model and thus makes it more difficult to solve.

Following this methodology, we can redefine the slack variables to finally introduce the logistic loss in the loss-optimal CT model discussed in Section \ref{sec:octs}:
\begin{equation}
\label{logistic_slacks}
\begin{split}
    \xi_{i,h}\ge f(v_k) + f'(v_k)(y_i(\boldsymbol{w}_t^T\boldsymbol{x}_i+b_t)-v_k)- M\Bigl(1 - \!\!\!\sum_{s \in \mathcal{T}_d(t)} \!\!\!z_{i, s} \Bigr),  \\ \forall \, i \in \mathcal{I},\, \forall\, h\in \mathcal{H},\,  \forall \, t \in \mathcal{T}_h,\;\forall v\in V.
\end{split}
\end{equation}

\subsection{Lasso Regularization}
\label{sec:l1}
Together with the loss function defining the slack values, the second element to be chosen in our generalized OCT framework is the regularizer. Using an $\ell_2$-regularization term, as for example in MARGOT, has the effect of making the entire problem an MIQP instance. Here, we instead propose to consider a Lasso regularizer \cite{tibshirani1996regression}, i.e., an $\ell_1$ penalty term; there are two main reasons for doing so: 
\begin{itemize}
    \item the $\ell_1$-norm can be easily handled by linear constraints within a linear programming model; thus, using the $\ell_1$-norm instead of the squared $\ell_2$-norm we can derive a fully linear model which should be easier to solve;
    \item  exploiting the well-known properties of the $\ell_1$-norm \cite{bach2012optimization}, we can implicitly induce sparsity within branch nodes classifiers, without the need of recurring to explicit (and expensive) $\ell_0$-norm penalization as in SFS models.
\end{itemize}

The $\ell_1$-norm can be efficiently handled in a linear program, similarly as what is done in \cite{figueiredo2007gradient}, setting
\begin{subequations}
\label{l1_in_lp}
    \begin{align}
        &\boldsymbol{w}_t=\boldsymbol{w}^+_t-\boldsymbol{w}^-_t\qquad\! \forall t\in \mathcal{T},\\
    & \boldsymbol{w}_t,\boldsymbol{w}^+_t,\boldsymbol{w}^-_t \in \mathbb{R}^p\qquad\!\!\!\!\!\forall t\in \mathcal{T},\\
    & w_{j, t}^+,w_{j, t}^- \geq 0 \qquad\quad\, \forall t \in \mathcal{T},\; \forall\, j=1,\ldots,p.
    \end{align}
\end{subequations}
Basically, $\boldsymbol{w}_t$ is split into its positive and negative parts $\boldsymbol{w}_t^+$ and $\boldsymbol{w}_t^-$. Indeed, constraints \eqref{l1_in_lp} are satisfied by $\hat{w}_{j,t}^+ = \max\{0,{w}_{j,t}\}$ and $\hat{w}_{j,t}^- = \max\{0,-{w}_{j,t}\}$; this solution is such that $\hat{w}_{j,t}^++\hat{w}_{j,t}^- = \lvert w_{j,t}\rvert$. We thus have  
$\boldsymbol{1}^T(\hat{\boldsymbol{w}}_t^++\hat{\boldsymbol{w}}_t^-) = \|\boldsymbol{w}_t\|_1$.

Any other feasible solution can be obtained by shifting $\hat{\boldsymbol{w}}_t^+$ and $\hat{\boldsymbol{w}}_t^-$ by a vector $\boldsymbol{\Delta}\ge 0$, $\|\boldsymbol{\Delta}\|>0$; hence, for any other feasible solution we have
\begin{align*}
    \boldsymbol{1}^T(\boldsymbol{w}^++\boldsymbol{w}^-) &= \boldsymbol{1}^T(\hat{\boldsymbol{w}}_t^++\boldsymbol{\Delta}+\hat{\boldsymbol{w}}_t^-+\boldsymbol{\Delta}) = \boldsymbol{1}^T(\hat{\boldsymbol{w}}^++\hat{\boldsymbol{w}}^-) + 2(\boldsymbol{1}^T\boldsymbol{\Delta})\\&> \boldsymbol{1}^T(\hat{\boldsymbol{w}}^++\hat{\boldsymbol{w}}^-) = \|\boldsymbol{w}_t\|_1. 
\end{align*}
The shift $\boldsymbol{\Delta}$, however, has no influence on the variables $\boldsymbol{w}_t$; if the term $\boldsymbol{1}^T(\boldsymbol{w}^++\boldsymbol{w}^-)$ is minimized in the objective, then $\hat{\boldsymbol{w}}_t^+$ and $\hat{\boldsymbol{w}}_t^-$ will always be chosen and thus the actual quantity being minimized is $\boldsymbol{1}^T(\hat{\boldsymbol{w}}^++\hat{\boldsymbol{w}}^-) = \|\boldsymbol{w}_t\|_1$. 

Lasso regularization is well known not only to induce sparsity and guarantee that the optimization process is well-behaved, but also to be beneficial at tackling overfitting \cite{hastie2009elements, tibshirani1996regression}. Thus, if the sparsity requirement within each node is not set by a specific budget (as in HFS), but it is imprecisely imposed by means of a penalty, then the expedient of setting $\Omega(\cdot) =\|\cdot\|_1$ allows to preserve, at a much lower cost, both the predictive performance of the obtained model and its interpretability.

The alternative SFS strategy from \cite{d2022margin} exploits the objective function 
$$\sum_{t\in \mathcal{T}}\Bigl(\Omega(\boldsymbol{w}_t) + C_t\sum_{i\in I} \xi_{i,t} + \alpha_t\max(\|\boldsymbol{w}_t\|_0, B_t)\Bigr).$$ 
This kind of penalty on the $\ell_0$-norm of weights is modeled with binary variables, see eq.\ \eqref{l0_bin}, highly increasing the complexity of the optimization model. Moreover, here there is an additional hyperparameter to be tuned for each layer, making the cross-validation procedure harder to manage.

\subsection{The Overall Model}
The overall \textit{Optimal Logistic Classification Tree} (OLCT) model proposed in this paper is obtained putting together the pieces described thus far:
\begin{subequations}
\label{logistic_tree_model}
    \begin{align}
        \min_{\substack{\boldsymbol{w},\boldsymbol{w}^+,\boldsymbol{w}^-\\ \boldsymbol{b},\boldsymbol{\xi}, \boldsymbol{z}}}\;&\sum_{h\in\mathcal{H}}\left(\sum_{i\in\mathcal{I}} \xi_{i,h}+\lambda_h\sum_{t\in\mathcal{T}_h}\|\boldsymbol{w}_t\|_1\right)\\\text{s.t. }&\eqref{z},\eqref{Forward},\eqref{logistic_slacks},\eqref{l1_in_lp} \label{all_constrs}\\& \boldsymbol{\xi_h}\in\mathbb{R}^n\quad \forall\, h\in\mathcal{H}, \label{def_all_slacks}\\ & b_t\in \mathbb{R}\qquad\forall\,t\in\mathcal{T}. \label{def_bias}
    \end{align}
\end{subequations}

The constraints referenced at \eqref{all_constrs} contain almost all the defining elements of the logistic tree: constraint \eqref{z} defines the indicator binary variables of data points routing, constraints \eqref{Forward} enforce the routing logic along the tree, constraints \eqref{logistic_slacks} define the slacks corresponding to the logistic loss and \eqref{l1_in_lp} define the branching classifiers weights and model their $\ell_1$-norm. 

Constraints \eqref{def_all_slacks} and \eqref{def_bias} simply define the domain of the remaining variables. The objective function is nothing but the general loss \eqref{gen_obj_full} where the $\ell_1$ regularizer is employed, formulated as shown in Section \ref{sec:l1}.

\begin{remark}
\label{remark:refine}
    Note that, in our model, we use a piece-wise linear underestimator, i.e.,  a surrogate function, to approximate the log-loss. Thus, at the end of the training process, we can actually perform a refinement operation affecting the last layer of branching nodes.
    Specifically, we can exactly fit an $\ell_1$-regularized logistic regression model at each node of the last layer, using as training data the samples that actually reach that node. By this procedure, we are guaranteed that the overall exact objective function associated with the entire tree model decreases. 

    The same reasoning cannot be applied with higher level nodes: changes to the branching classifiers possibly change how training data are forwarded to lower nodes, thus affecting the corresponding loss terms. The overall loss associated with the tree might increase, because of the lack of global perspective.
\end{remark}

\begin{remark}
    One of the most appealing features of classification trees is their glass-box nature; however, the highest level of interpretability is only reached in the case of parallel splits. For this reason, we believe it to be important to explicitly point out how to retrieve a univariate optimal logistic classification trees. The model is basically equivalent to \eqref{logistic_tree_model}, except for HFS type constraints, with an upper bound on the $\ell_0$-norm of each branching classifier set to 1. This can be modeled in MILP terms by constraints \eqref{l0_bin} and setting $\boldsymbol{1}^T\boldsymbol{\delta}_t\le 1$.

Of course, the addition of a number of new binary variables and big-$M$ type constraints directly proportional to $p\times \lvert\mathcal{T}\rvert$ is significant in terms of the computational resources needed to solve the problem and especially to certify optimality, closing the optimality gap.     
\end{remark}


\subsection{Interpreting OLCTs}
\label{subseq:interpreting}
Inheriting, at least partially, the nice interpretability properties of logistic regression models is one of the main advantages of using the logistic loss within the OCTs framework.

Specifically, there are some aspects that can be taken into account to retrieve additional information about the model prediction mechanisms.
\begin{description}
    \item[Evaluation of feature influence] At each branching node of oblique OLCTs, the influence of each individual parameter in the splitting decision can be estimated looking at the magnitude of the corresponding coefficient and multiplying it by the standard deviation of the feature among the training data points reaching that node; formally, given the weights $\boldsymbol{w}_t$ at a node $\bar{t}\in\mathcal{T}$, the influence $r_{j,\bar{t}}$ of feature $j$ can be estimated by:
\begin{gather*}
    N_{\bar{t}} = \sum_{i\in \mathcal{I}}\sum_{t\in\mathcal{T}_d(t)}z_{i,t}, \qquad \mu_{j,\bar{t}} = \frac{1}{N_{\bar{t}}}\sum_{i\in\mathcal{I}}\left(x_{i,j}\sum_{t\in\mathcal{T}_d(\bar{t})}z_{i,t}\right),\\  \sigma_{j,\bar{t}} = \sqrt{\frac{1}{N_{\bar{t}}}\sum_{i\in \mathcal{I}}\left((x_{i,j}-\mu_{j,\bar{t}})^2\sum_{t\in\mathcal{T}_d(t)}z_{i,t}\right)},\qquad r_{j,\bar{t}} = w_{j,\bar{t}}\sigma_{j,\bar{t}}.
\end{gather*}
The larger the coefficient $r_{j,t}$ is, the higher is the connection of feature $j$ with positive outputs; on the other hand, the largest negative values of importance are associated with features pushing the point towards the left child of the branching node. The features influence for an OLCT trained on the \texttt{heart} dataset is reported, as an example, in Figure \ref{fig:importance}.

\begin{figure}[h]
    \centering

\tikzset{every picture/.style={line width=0.75pt}} 

\begin{tikzpicture}[x=0.75pt,y=0.75pt,yscale=-1,xscale=1]

\draw    (304.42,124.25) -- (201.42,182.75) ;
\draw    (328.42,125.75) -- (432.42,184.75) ;
\draw    (434.42,203.75) -- (383.42,254.75) ;
\draw    (456.42,202.75) -- (505.42,253.75) ;
\draw  [dash pattern={on 4.5pt off 4.5pt}]  (383.42,52.25) -- (321.42,103.25) ;
\draw  [dash pattern={on 4.5pt off 4.5pt}]  (519.42,154.25) -- (457.42,184.25) ;

\draw  [fill={rgb, 255:red, 255; green, 255; blue, 255 }  ,fill opacity=1 ]  (316.23, 117.2) circle [x radius= 13.6, y radius= 13.6]   ;
\draw (316.23,117.2) node    {$0$};
\draw  [fill={rgb, 255:red, 155; green, 155; blue, 155 }  ,fill opacity=1 ]  (190.23, 191.2) circle [x radius= 13.6, y radius= 13.6]   ;
\draw (190.23,191.2) node    {$1$};
\draw  [fill={rgb, 255:red, 255; green, 255; blue, 255 }  ,fill opacity=1 ]  (445.23, 193.2) circle [x radius= 13.6, y radius= 13.6]   ;
\draw (445.23,193.2) node    {$2$};
\draw  [fill={rgb, 255:red, 155; green, 155; blue, 155 }  ,fill opacity=1 ]  (376.23, 267.2) circle [x radius= 13.6, y radius= 13.6]   ;
\draw (376.23,267.2) node    {$3$};
\draw  [fill={rgb, 255:red, 155; green, 155; blue, 155 }  ,fill opacity=1 ]  (512.23, 267.2) circle [x radius= 13.6, y radius= 13.6]   ;
\draw (512.23,267.2) node    {$4$};
\draw (162,209.4) node [anchor=north west][inner sep=0.75pt]    {$-1$};
\draw (349,286.4) node [anchor=north west][inner sep=0.75pt]    {$-1$};
\draw (521,288.4) node [anchor=north west][inner sep=0.75pt]    {$1$};
\draw (273,107.4) node [anchor=north west][inner sep=0.75pt]    {$\leq 0$};
\draw (400,189.4) node [anchor=north west][inner sep=0.75pt]    {$\leq 0$};
\draw (334,107.4) node [anchor=north west][inner sep=0.75pt]    {$ >0$};
\draw (463,189.4) node [anchor=north west][inner sep=0.75pt]    {$ >0$};
\draw    (385,19) -- (464,19) -- (464,70) -- (385,70) -- cycle  ;
\draw (388,21.4) node [anchor=north west][inner sep=0.75pt]  [font=\footnotesize]  {$ \begin{array}{l}
r_{0,3} :4.20\\
r_{0,7} :-10.69\\
r_{0,9} :0.4\\
r_{0,11} :1.1
\end{array}$};
\draw    (520,136) -- (583,136) -- (583,165) -- (520,165) -- cycle  ;
\draw (523,138.4) node [anchor=north west][inner sep=0.75pt]  [font=\footnotesize]  {$ \begin{array}{l}
r_{2,2} :0.09\\
r_{2,10} :0.03
\end{array}$};

\node[] () [below = 1.5em] at (350,285) {};
\end{tikzpicture}

    \caption{A learned logistic oblique tree of depth 2 for the \texttt{heart} dataset. Branches and leaves are on white and gray background respectively. Each branch node is a sparse linear regressor; the importance coefficient for each feature involved at each decision is reported in the boxes.}
    \label{fig:importance}
\end{figure}
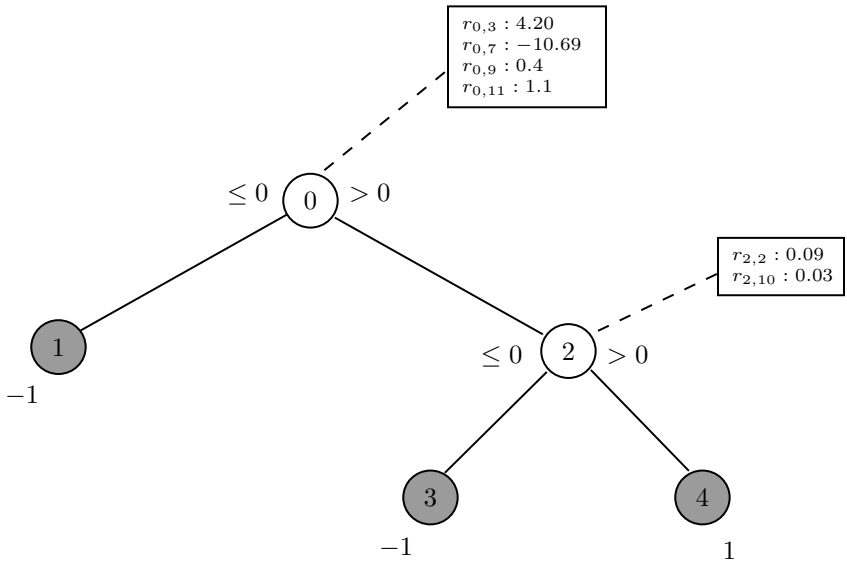


\item[Probabilistic interpretation of outputs and model calibration]

The sigmoid function $\sigma(z) = \frac{1}{1+\exp(-z)}$ is used in logistic regression to map the output of the linear function defined by $\boldsymbol{w},b$ to $(0,1)$, so that a data point $\boldsymbol{x}$ can finally be classified as positive if $\sigma(\boldsymbol{w}^T\boldsymbol{x}+b)\ge 0.5$.

In a standalone logistic regression model, these ``probability values'' are related to the odds of the positive outcome over the negative one; in particular, the linear regressor is designed to model, by maximum likelihood, the log-odds (logits) of the output:
$$\boldsymbol{w}^T\boldsymbol{x}+b = \text{logit}(\mathbb{P}(y=1\mid \boldsymbol{x})) = \log \left(\frac{\mathbb{P}(y=1\mid \boldsymbol{x})}{\mathbb{P}(y=-1\mid \boldsymbol{x})}\right).$$
Exponentiating we get
$$\exp(\boldsymbol{w}^T\boldsymbol{x}+b)=\frac{\mathbb{P}(y=1\mid \boldsymbol{x}))}{1-\mathbb{P}(y=1\mid \boldsymbol{x}))}, $$
and by simple algebraic manipulations we retrieve
$$\mathbb{P}(y=1\mid \boldsymbol{x}) = \frac{\exp(\boldsymbol{w}^T\boldsymbol{x}+b)}{1+\exp(\boldsymbol{w}^T\boldsymbol{x}+b)} = \frac{1}{1+\exp(-\boldsymbol{w}^T\boldsymbol{x}-b)}=\sigma(\boldsymbol{w}^T\boldsymbol{x}+b).$$
In other words, the logistic output is an actual probability estimate for the positive output, i.e., the logistic model is well \textit{calibrated} by construction.

Within the OLCTs framework, we can exploit this property both at the splitting and the classifying nodes of the tree.
If $t\in\mathcal{T}_d$, then the odds associated with the outputs of $\boldsymbol{w}_t, \ b_t$ are actually interpretable as the odds of the positive class over the negative one, given that the data point belongs to the subspace deterministically defined by the splits at the higher nodes. 
Since each node classifier is calibrated within the corresponding space region, the overall output probability of the OLCT models should also be implicitly well-calibrated, with no need of any extra post-train calibration. 

If, on the other hand, $t\in\mathcal{T}_h$, $h<d$, then the classifier has been trained looking not only at its corresponding slacks, but also at those of all its descendant nodes. Thus, in this case we can (somewhat improperly) interpret the probability estimates as the confidence about forwarding the data point to the right child rather than to the left one. This concepts can be visualized through the example in Figure \ref{fig:Tree3}.

\begin{figure}[h]
    \centering
    \tikzset{every picture/.style={line width=0.75pt}} 

\begin{tikzpicture}[x=0.75pt,y=0.75pt,yscale=-1,xscale=1]

\draw   (300.75,83.13) .. controls (300.75,74.77) and (307.52,68) .. (315.88,68) .. controls (324.23,68) and (331,74.77) .. (331,83.13) .. controls (331,91.48) and (324.23,98.25) .. (315.88,98.25) .. controls (307.52,98.25) and (300.75,91.48) .. (300.75,83.13) -- cycle ;

\draw    (315.42,37.25) -- (315.85,66) ;
\draw [shift={(315.88,68)}, rotate = 269.15] [color={rgb, 255:red, 0; green, 0; blue, 0 }  ][line width=0.75]    (10.93,-3.29) .. controls (6.95,-1.4) and (3.31,-0.3) .. (0,0) .. controls (3.31,0.3) and (6.95,1.4) .. (10.93,3.29)   ;
\draw  [fill={rgb, 255:red, 155; green, 155; blue, 155 }  ,fill opacity=1 ] (174.75,158.13) .. controls (174.75,149.77) and (181.52,143) .. (189.88,143) .. controls (198.23,143) and (205,149.77) .. (205,158.13) .. controls (205,166.48) and (198.23,173.25) .. (189.88,173.25) .. controls (181.52,173.25) and (174.75,166.48) .. (174.75,158.13) -- cycle ;
\draw   (430.75,159.13) .. controls (430.75,150.77) and (437.52,144) .. (445.88,144) .. controls (454.23,144) and (461,150.77) .. (461,159.13) .. controls (461,167.48) and (454.23,174.25) .. (445.88,174.25) .. controls (437.52,174.25) and (430.75,167.48) .. (430.75,159.13) -- cycle ;
\draw    (303.42,92.75) -- (201.42,148.75) ;
\draw    (328.42,91.75) -- (432.42,150.75) ;
\draw  [fill={rgb, 255:red, 155; green, 155; blue, 155 }  ,fill opacity=1 ] (358.75,232.13) .. controls (358.75,223.77) and (365.52,217) .. (373.88,217) .. controls (382.23,217) and (389,223.77) .. (389,232.13) .. controls (389,240.48) and (382.23,247.25) .. (373.88,247.25) .. controls (365.52,247.25) and (358.75,240.48) .. (358.75,232.13) -- cycle ;
\draw  [fill={rgb, 255:red, 155; green, 155; blue, 155 }  ,fill opacity=1 ] (500.75,232.13) .. controls (500.75,223.77) and (507.52,217) .. (515.88,217) .. controls (524.23,217) and (531,223.77) .. (531,232.13) .. controls (531,240.48) and (524.23,247.25) .. (515.88,247.25) .. controls (507.52,247.25) and (500.75,240.48) .. (500.75,232.13) -- cycle ;
\draw    (434.42,169.75) -- (383.42,220.75) ;
\draw    (457.42,169.75) -- (506.42,220.75) ;
\draw  [dash pattern={on 4.5pt off 4.5pt}]  (286.42,86.75) -- (209.18,127.81) ;
\draw [shift={(207.42,128.75)}, rotate = 332] [color={rgb, 255:red, 0; green, 0; blue, 0 }  ][line width=0.75]    (10.93,-3.29) .. controls (6.95,-1.4) and (3.31,-0.3) .. (0,0) .. controls (3.31,0.3) and (6.95,1.4) .. (10.93,3.29)   ;
\draw [line width=1.5]    (345.42,85.75) -- (425.83,133.22) ;
\draw [shift={(428.42,134.75)}, rotate = 210.56] [color={rgb, 255:red, 0; green, 0; blue, 0 }  ][line width=1.5]    (14.21,-4.28) .. controls (9.04,-1.82) and (4.3,-0.39) .. (0,0) .. controls (4.3,0.39) and (9.04,1.82) .. (14.21,4.28)   ;
\draw [line width=1.5]    (467.42,163.75) -- (506.35,204.58) ;
\draw [shift={(508.42,206.75)}, rotate = 226.36] [color={rgb, 255:red, 0; green, 0; blue, 0 }  ][line width=1.5]    (14.21,-4.28) .. controls (9.04,-1.82) and (4.3,-0.39) .. (0,0) .. controls (4.3,0.39) and (9.04,1.82) .. (14.21,4.28)   ;
\draw  [dash pattern={on 4.5pt off 4.5pt}]  (425.42,167.75) -- (385.92,202.43) ;
\draw [shift={(384.42,203.75)}, rotate = 318.72] [color={rgb, 255:red, 0; green, 0; blue, 0 }  ][line width=0.75]    (10.93,-3.29) .. controls (6.95,-1.4) and (3.31,-0.3) .. (0,0) .. controls (3.31,0.3) and (6.95,1.4) .. (10.93,3.29)   ;

\draw (311,77) node [anchor=north west][inner sep=0.75pt]    {$0$};
\draw (314,20) node     {$\boldsymbol{x}$};
\draw (441,153) node [anchor=north west][inner sep=0.75pt]    {$2$};
\draw (185,152) node [anchor=north west][inner sep=0.75pt]    {$1$};
\draw (369,225) node [anchor=north west][inner sep=0.75pt]    {$3$};
\draw (511,225) node [anchor=north west][inner sep=0.75pt]    {$4$};
\draw (156,182.4) node [anchor=north west][inner sep=0.75pt]    {$-1$};
\draw (351,253.4) node [anchor=north west][inner sep=0.75pt]    {$-1$};
\draw (523,254.4) node [anchor=north west][inner sep=0.75pt]    {$1$};
\draw (210,82.4) node [anchor=north west][inner sep=0.75pt]    {$p=0.3$};
\draw (377,82.4) node [anchor=north west][inner sep=0.75pt]    {$p=0.7$};
\draw (491,161.4) node [anchor=north west][inner sep=0.75pt]    {$p=0.81$};
\draw (350,166.4) node [anchor=north west][inner sep=0.75pt]    {$p=0.19$};

\node[] () [below = 1.5em] at (350,260) {};
\end{tikzpicture}

\caption{Confidence at each branch node of the logistic classification tree from Figure \ref{fig:importance} for the sample $\boldsymbol{x} = [ 0.69,  0.87, -1.23, -0.8, -0.4, 1.01, -1.87,  1.4  , -0.94,  0.63, 0.35, -0.89, -0.07]$ of the \texttt{heart} dataset. The true label $y$ is equal to 1. In this case, the model predicts the correct class of the point and, given the sigmoid activation, we are able to get the confidence of the forwarding decision at each branch node.}
\label{fig:Tree3}

\end{figure}
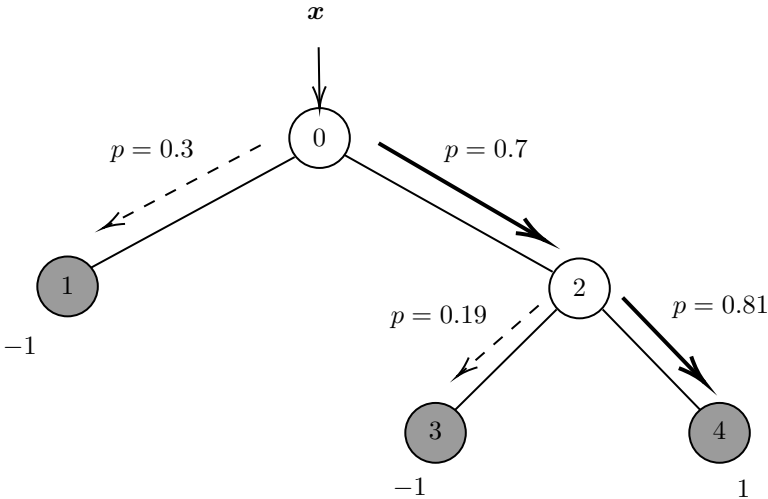



\end{description}

%
  


\section{Numerical Experiments}
\label{sec:exp}
In this section, we present the results of computational experiments carried out to evaluate the performance of the proposed approach. All the experiments described in this section have been carried out on a server with an Intel \textregistered  Xeon \textregistered  Gold 6330N CPU with 28 cores and 56 threads @ 2.20GHz, but we set a limit of only 40 of the 56 available threads and the total available memory is 128GB. The code has been implemented in Python (v.\ 3.9) and the commercial solver Gurobi \cite{gurobi} has been used to solve all the mixed-integer programming models considered in this work. All the code is available at \url{https://github.com/tom1092/Optimal-Logistic-Classification-Trees}.

For each instance of classification tasks, we performed an 80/20 train/test split of the data and we also standardized each feature before the training to zero mean and unit variance. Experiments are always repeated for different random seeds, resulting in different train/test splits. Hyperparameters have been tuned by cross-validation over a grid of values, where the test balanced accuracy (see below) is used as quality metric; more details will be provided for each group of experiments in the following. The parameters $M$ and $\epsilon$ in MIP formulations have been set to 100 and $10^{-5}$ respectively; the value of $M=100$ is a reasonable value, large enough not to introduce bound constraints and tight enough not to make Gurobi branch-and-bound too inefficient; this same value was also used, for instance, in \cite{d2022margin}; as for $\epsilon$, the value $10^{-5}$ is much larger than the one employed for the IntFeasTol parameter of Gurobi ($10^{-9}$), so that the issues highlighted in \cite{liu2023optimal} did never occur, and at the same time is small enough to let the underlying logic of the constraints work properly - no feasible solution is erroneously cut off.  Without loss of generality, we also decided to move the regularization parameter $\lambda_h$ to the slack component of the loss, to be aligned with the work in \cite{d2022margin}. The objective function now has the form:
$$\sum_{h\in\mathcal{H}}\left(C_h\sum_{i\in\mathcal{I}} \xi_{i,h}+\sum_{t\in\mathcal{T}_h}\|\boldsymbol{w}_t\|_1\right)$$

Note that, at the end of the optimization process of OLCTs,  we applied the refinement strategy discussed in Remark \ref{remark:refine}. For each node in the last layer, we retrained the $\ell_1$-regularized logistic regression model using \texttt{scikit-learn} \cite{scikit-learn} implementation. 

To compare the performance of each model, along with the running times, we used the balanced accuracy metric defined as:
$$\text{B}_\text{Acc} = \frac{\text{Sensitivity} + \text{Specificity}}{2} = \frac{1}{2} \left( \frac{TP}{TP+FN}+ \frac{TN}{TN+FP}\right),$$
where TP, TN, FP, FN are the number of true positive, true negative, false positive and false negative outputs, respectively. The B$_\text{Acc}$ value is always computed on test set data.

To provide a condensed view of the results, in the following we are making use of performance profiles \citep{dolan2002benchmarking}.
Performance profiles provide a unified view of the relative performance of the solvers on a suite of test problems. 
Formally, consider a benchmark of $\mathcal{P}$ problem instances and a set of solvers $\mathcal{S}$. For each solver $\sigma\in\mathcal{S}$ and problem $\pi\in\mathcal{P}$, we define
$$c_{\pi,\sigma} =  \text{the cost for solver $\sigma$ to solve problem } \pi,$$
where cost is the performance metric we are interested in. In particular, we will be interested in CPU time.
We then consider the ratio
$$\eta_{\pi,\sigma} = \frac{c_{\pi,\sigma}}{\min_{\sigma\in \mathcal{S}}\{c_{\pi, \sigma}\}},$$
which expresses a relative measure of the performance on problem $\pi$
of solver $\sigma$ against the performance of the best solver for this problem. If a solver fails to solve a problem, we shall put $\eta_{\pi, \sigma} = \eta_M$, with $\eta_M \geq \max\{\eta_{\pi, \sigma}\mid \pi\in\mathcal{P},\,\sigma\in\mathcal{S} \}$.

Finally, the performance profile for a solver $\sigma$ is given by the function
$$\rho_\sigma(\tau) = \frac{1}{\lvert\mathcal{P}\rvert} \cdot \left\lvert\left\{\pi \in \mathcal{P} \mid \eta_{\pi,\sigma} \leq \tau\right\}\right\rvert,$$
which represents the estimated probability for solver $\sigma$ that the performance ratio $\eta_{\pi,\sigma}$ on an arbitrary instance $\pi$ is at most $\tau \in \mathbb{R}$. The function $\rho_\sigma(\tau):[1, +\infty]\to [0, 1]$ is, in fact, the cumulative distribution of the performance ratio.

Note that the value of $\rho_\sigma(1)$ is the fraction of problems where solver $\sigma$ attained the best performance; on the other hand, $\lim_{\tau \to \eta_M^-} \rho_\sigma(\tau)$ denotes the fraction of problems solved from the given benchmark.

\subsection{Preliminary experiments}
\label{sec:prelim}
The first experiments we carried out concern the assessment of the performance of our model as we vary the set $V$ of the tangent points that are used to obtain the piece-wise linear underestimator of the logistic loss.
In particular, we are interested in finding a good trade-off between the quality of the approximation and the running time of the model.

We considered a small benchmark of 5 datasets (\texttt{parkinsons, wholesale, tik-tak-toe, haberman, sonar}, see Table \ref{table:dataset}) from the UCI repositories \cite{Dua:2019}, testing $V_0, V_1, V_2$ with refinment as configurations for the MILP problem of training an OLCT of depth 2. The experiment has been repeated for five different random seeds for each dataset for a total of 25 different problems.  The Gurobi time limit has been set to 300s. In these experiments, we did not employ the $\ell_1$ regularization terms, as we are mainly interested in assessing the effectiveness of log loss approximation. Note that the MIP solution process is warm-started, initializing the weights of each branch node following a strategy similar to the one adopted in \cite{d2022margin}: starting from the root in a greedy fashion, we assign to each node weights the values obtained training a logistic regression classifier using the data that are forwarded to that node by the above branches. This strategy allows to obtain significant speedups in computation.

\begin{figure}[h]
    \centering
    \begin{subfigure}[t]{0.48\textwidth}
        \includegraphics[width=\textwidth]{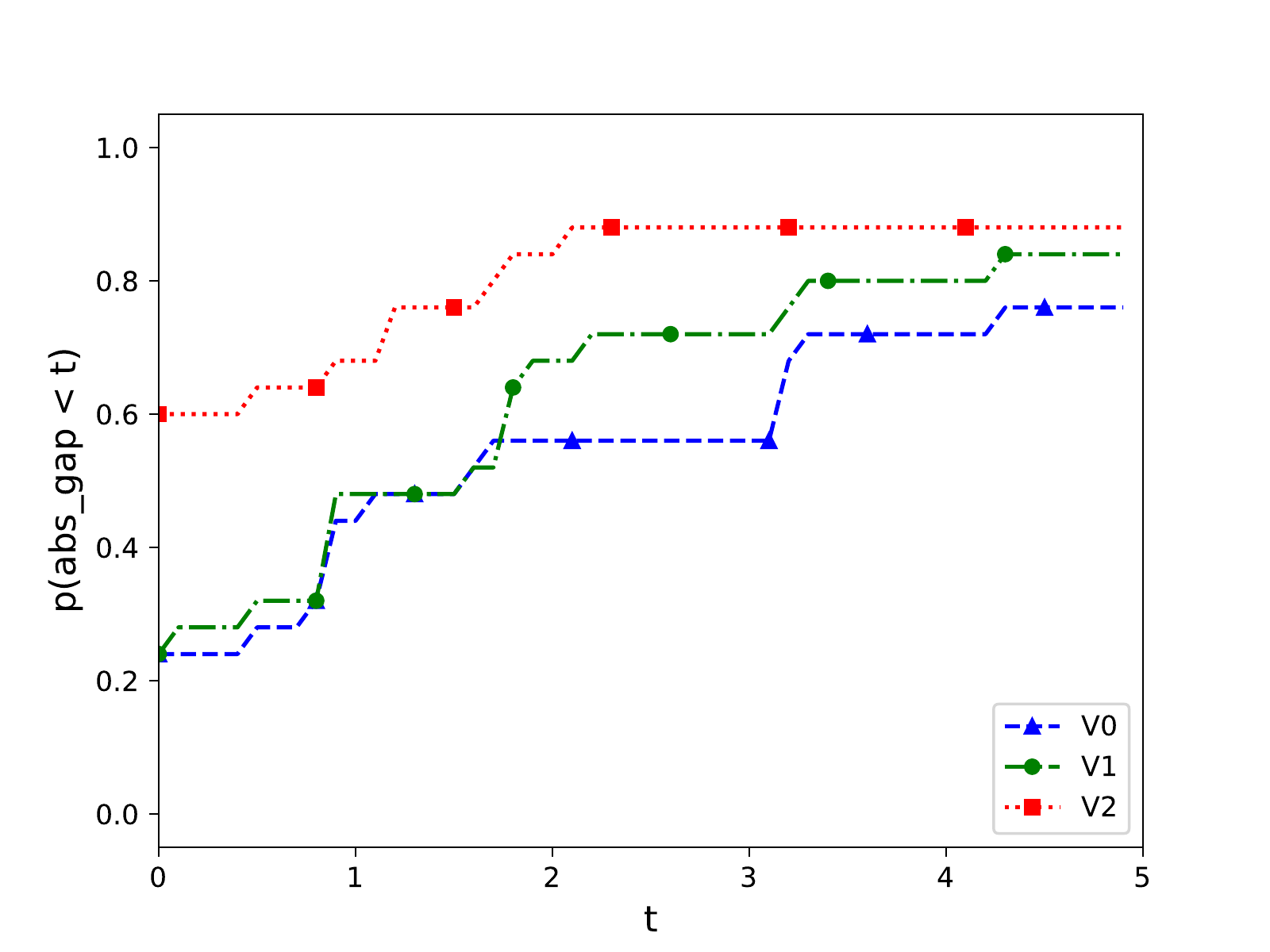}
        \caption{Cumulative distribution of the absolute gap from the best test balanced accuracy value attained by any model.}
    \label{subfig:baccuracy}
    \end{subfigure}
    \hfill
    \begin{subfigure}[t]{0.48\textwidth}
        \includegraphics[width=\textwidth]{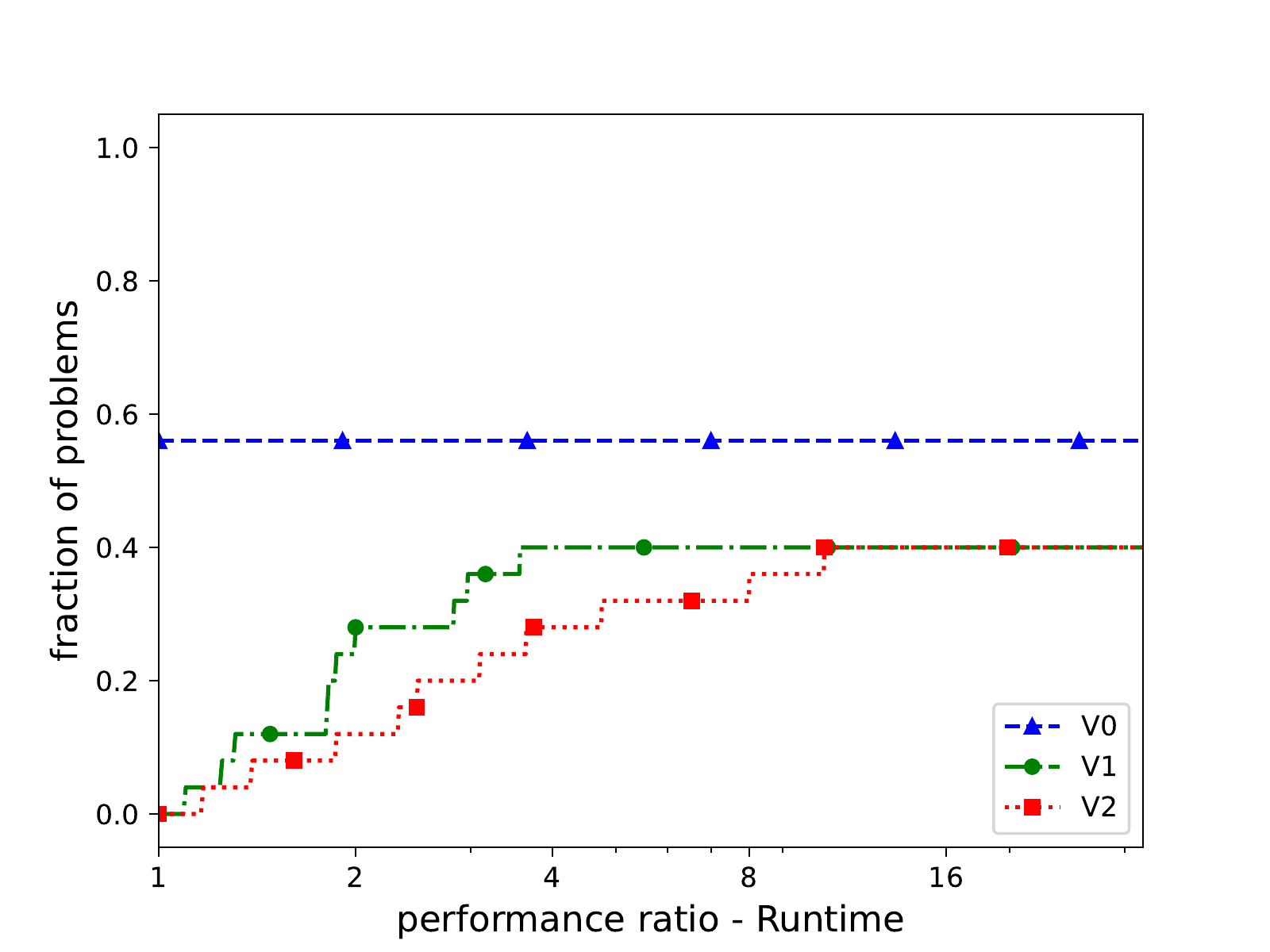}
        \caption{Performance profiles of the running times for solving the MIP models.}
    \label{subfig:time}
    \end{subfigure}
    \begin{subfigure}[t]{0.48\textwidth}
        \includegraphics[width=\textwidth]{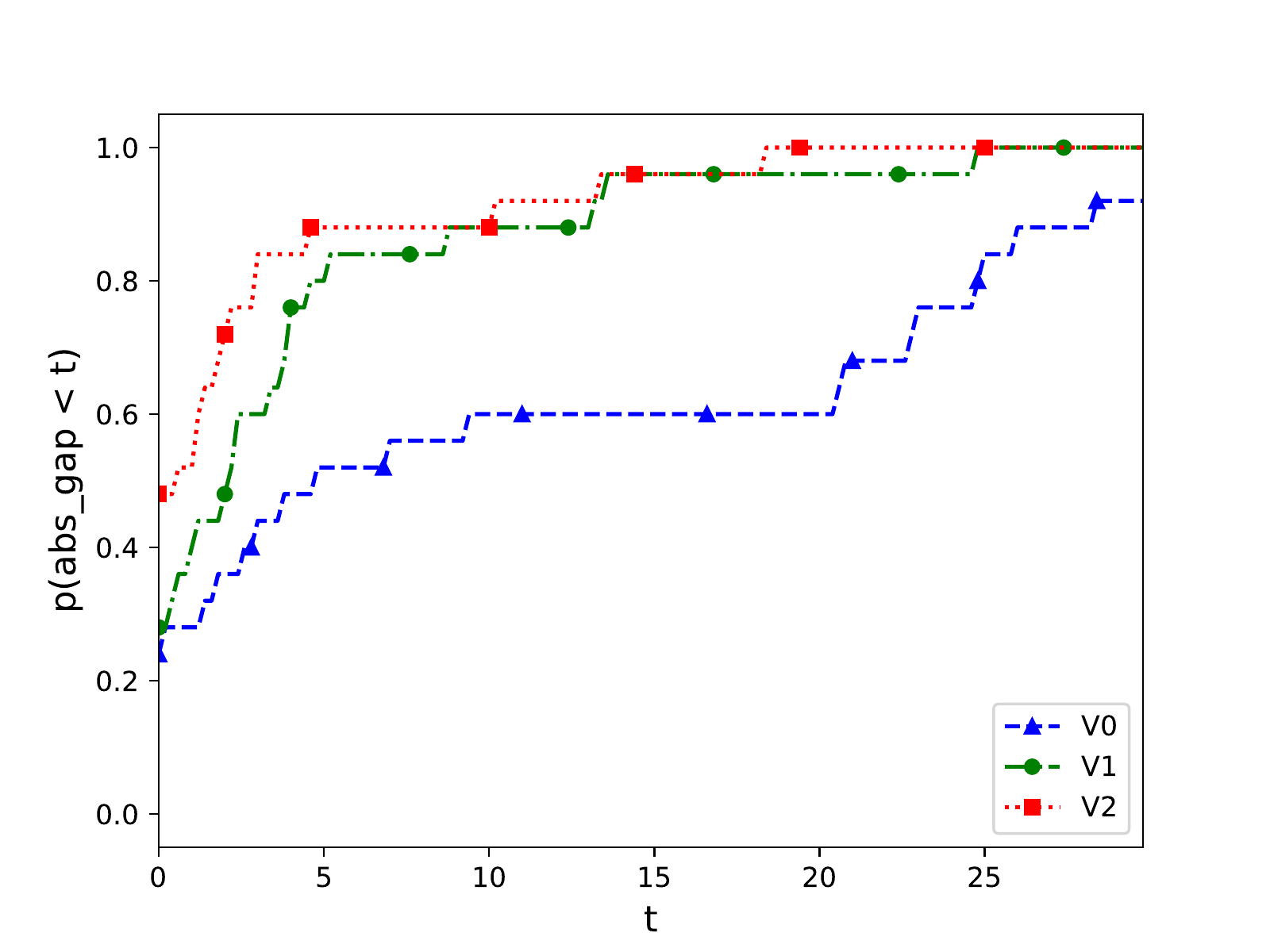}
        \caption{Cumulative distribution of the absolute gap from the best (exact) loss attained on the training set by any model.}
    \label{subfig:sparsity}
    \end{subfigure}
    \caption{Comparison of the performance of OLCT models when different tangent point sets ($V_0$, $V_1$ or $V_2$) are employed. At the end of the optimization process, the refinement of the last layer was carried out for each model.}
    \label{fig:results-vs}
\end{figure}

As shown by the performance profiles in Figure \ref{fig:results-vs}, choosing $V_0=\{ 0, \pm \infty\}$ to build the linear piece-wise approximation seems to provide a nice trade-off between running time and solution quality. Indeed, the use of $V_1$ does not seem to significantly improve the out-of-sample accuracy of the model; on the other hand, the more accurate approximation obtained with $V_2$ does result in better predictive models, but a much higher training cost has to be paid.  
For this reason, we decided to use the $V_0$ setting to assess the performance of our model in the following sections. Nonetheless, we do not rule out that, in certain settings, it might be worth exploiting the increased effectiveness provided by $V_2$.  

\subsection{Impact Assessment for the Global Optimization Approach}
\begin{figure}[h]
    \centering
    \begin{subfigure}[t]{0.48\textwidth}
        \includegraphics[width=\textwidth]{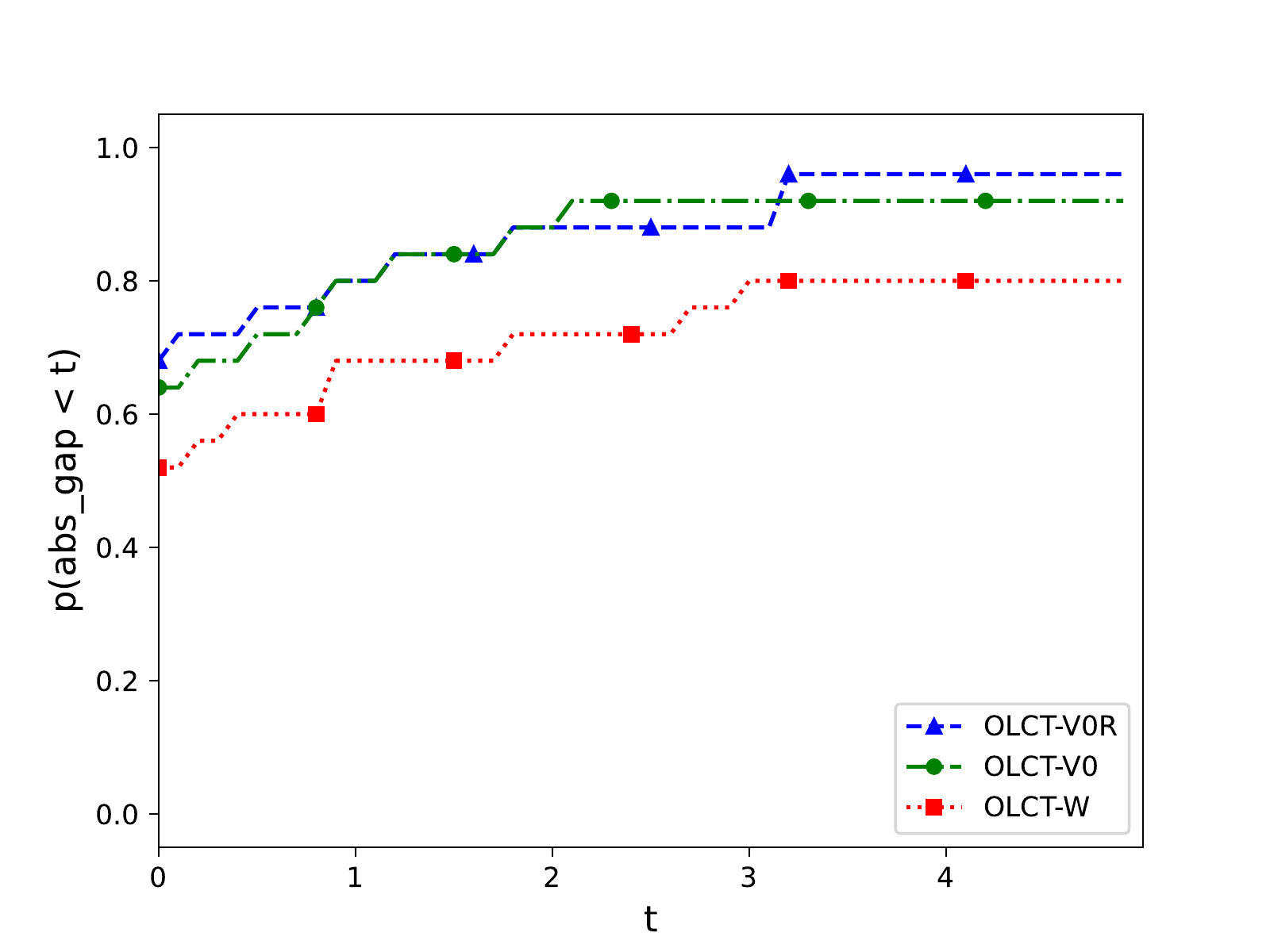}
        \caption{Cumulative distribution of the absolute gap from the best test balanced accuracy value attained by any model.}
    \label{subfig:baccuracy-go}
    \end{subfigure}
    \hfill
    \begin{subfigure}[t]{0.48\textwidth}
        \includegraphics[width=\textwidth]{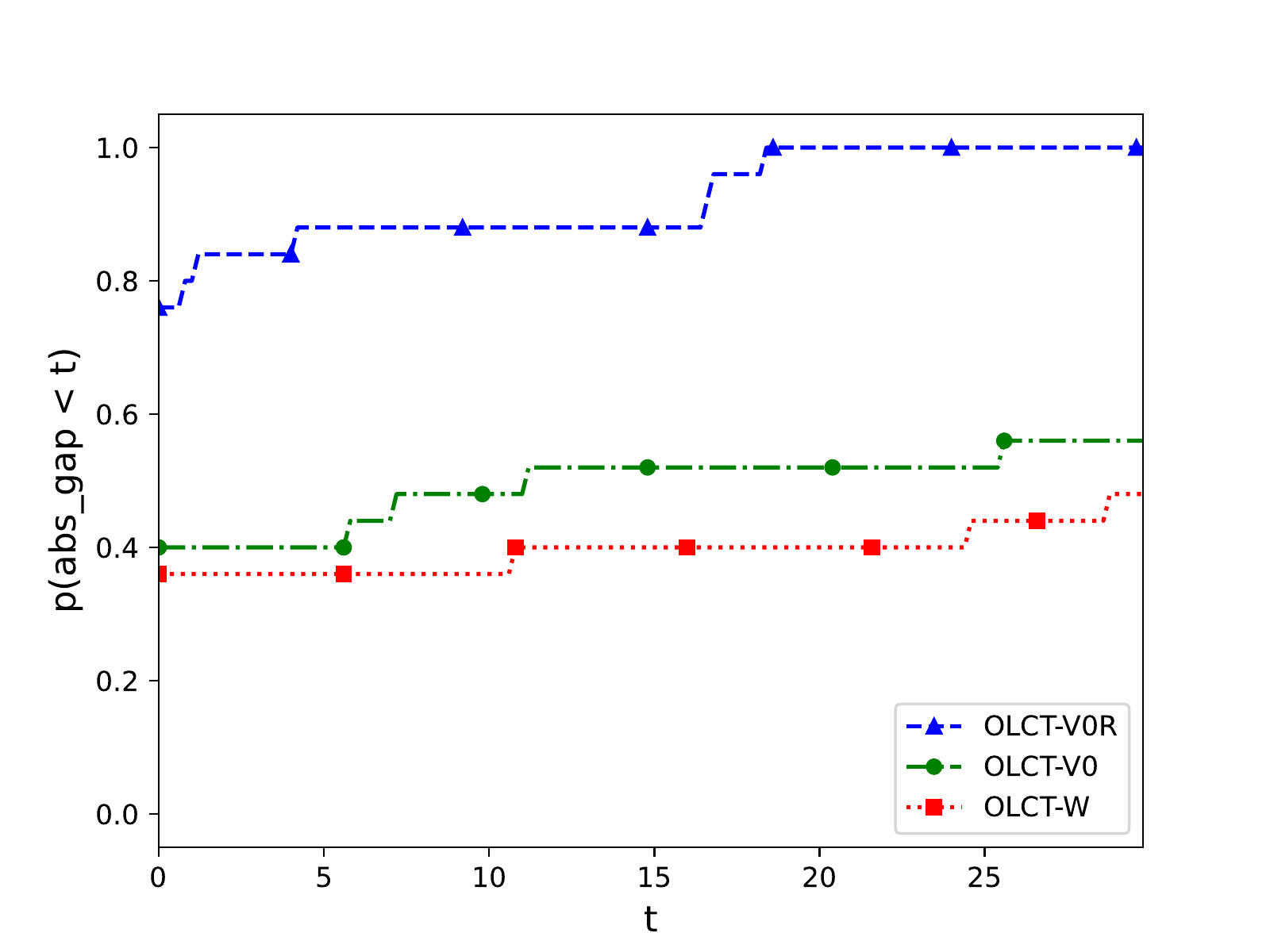}
        \caption{Cumulative distribution of the absolute gap from the best (exact) loss attained on the training set by any model.}
    \label{subfig:trueloss-go}
    \end{subfigure}
    \caption{Performance of logistic CTs (test balanced accuracy and train logistic loss) obtained carrying out: only the warm start procedure (OLCT-W); warm-start and global optimization with $V_0$ approximation (OLCT-V0); warm start, global optimization with $V_0$ and last branch exact refinement (OLCT-V0R).}
    \label{fig:results-go}
\end{figure}

In this Section, we investigate the actual beneficial effects of conducting a global optimization phase during logistic trees training. Indeed, good performance of OLCTs might be mainly due, in principle, to the warm-start or the refinement steps.
In particular, our greedy warm-start procedure constructs a logistic tree model with an iterative top-down approach, which is somewhat similar to the one proposed in \cite{chan2004lotus} and therefore might actually already lead to a properly effective classifier. 
However, the results reported in Figure \ref{fig:results-go} suggest that solving model \eqref{logistic_tree_model} does provide a substantial boost to the performance of the resulting logistic tree.

In this experiment, we examined the value of the overall (in-sample) logistic loss associated with the model after the warm-start, global optimization and refinement steps, and then we also took into account the (out-of-sample) values of balanced accuracy.
Here, we solved the same 25 instances of classification problems considered in Section \ref{sec:prelim}. Again, in order to focus on the consequences of approximating the loss in the global step, we did not employ the $\ell_1$ regularization terms. We also set Gurobi time limit to 300s. Moreover, based on the results in the previous section, we chose to employ the set $V_0$ of tangent points in log loss approximation. We did not focus on the running times of the three phases, as the global optimization step obviously represents the main computational burden for our approach.

From Figure \ref{subfig:trueloss-go}, we observe that solving the training problem with a global structure perspective generally allows to slightly improve the overall loss attained by the greedy model. This result has to be underlined, taking into account that the loss function is roughly approximated during the global optimization phase. Then, the refinement step on the last layer allows to really polish the model on training data, often leading to substantially lower values of loss.

The results in Figure \ref{subfig:baccuracy-go} are even more appealing. Apparently, when it comes to test performance, learning branching rules with a global point of view is crucial to improve the effectiveness of the resulting model. In this perspective, although visible, the positive effect of refinement is much more limited. Thus, we can arguably state that the solving \eqref{logistic_tree_model} as a global optimization has a significant effect in improving the effectiveness of logisitic tree models.

\subsection{OLCTs Performance Evaluation}
We now present the results of a larger computational experiment where we compare the OLCT model to the MARGOT, SFS-MARGOT \cite{d2022margin} and OCT-H \cite{bertsimas2017optimal} approaches. To assess the performance of our method, we considered 10 standard binary classification datasets from the UCI repositories \cite{Dua:2019} that are reported in Table \ref{table:dataset}. For each dataset, we consider 5 classification problem instances obtained considering different random train/test splits, so that the overall benchmark is made up of 50 test instances. For each problem instance, we set the depth of each tree model to 2. For all models, we set the Gurobi time limit to 300s both for validation runs and the final fit over the entire training set.

\begin{table}[htb]
\centering
\begin{tabular}{lcc}
\hline
        \textbf{Dataset} & \textbf{$n$} & \textbf{$p$} \\
        \hline
        breast & 568 & 30 \\
        climate & 539 & 18 \\
        haberman & 305 & 13 \\
        heart & 296 & 13 \\
        ionosphere & 350 & 33 \\
        parkinsons & 194 & 22 \\
        sonar & 207 & 60 \\
        spectf & 266 & 44 \\
        tik-tak-toe & 957 & 27 \\
        wholesale & 439 & 7 \\\hline
    \end{tabular}
    \vspace{1em}
    \caption{Description of the datasets used in the computational experiments. All datasets are from the UCI collection \cite{Dua:2019}.}
    \label{table:dataset}
\end{table}

For hyperparameters tuning, this time we used a 4-fold cross-validation. For both OLCT and MARGOT, we considered the slack parameters $C_h\in\{ 10^i, \ i = -2, -1, ..., 2\} \ \ \forall \ h \in \mathcal{H}$, obtaining 25 possible model configurations. On the other hand, the SFS-MARGOT model has two hyperparameters for the classifiers at each level: the slack parameter $C_h$ and the $\ell_0$ penalty parameter $\alpha_h$. In order to consider a comparable grid of configurations in size as that of OLCTs and MARGOTs, we used the same $\alpha$ for each branch layer $h$ letting $C_h$ vary in $\{10^{-2}, 1, 10^{2}\}$ and $\alpha$ in $\{10^{-1}, 1, 10\}$, so that a total number of 27 models is considered for the SFS variant. Finally for OCT-H we used the grid $\alpha \in\{ 2^i, \ i = -8, -1, ..., 2\} \cup \{0\} $ to tune the $\alpha$ parameter that penalizes the number of features used at each branch node to make the decision.

Again, we initialize the training phase of each MIP model by injecting a \textit{warm start} solution, obtained training logistic or SVM classifiers, depending on the particular tree classifier. For OCT models we used an analogous greedy strategy: we solve the MILP problem at each individual single node, i.e., setting the depth of the tree equal to 1 and only using the set of points reaching the considered branch node, with a time limit of 30s. As also mentioned in \cite{bertsimas2017optimal}, this strategy can significantly speedup the optimization, providing a good initial upper bound of the loss that may help the branch and bound method.

\begin{figure}[h]
    \centering
    \begin{subfigure}[t]{0.48\textwidth}
        \includegraphics[width=\textwidth]{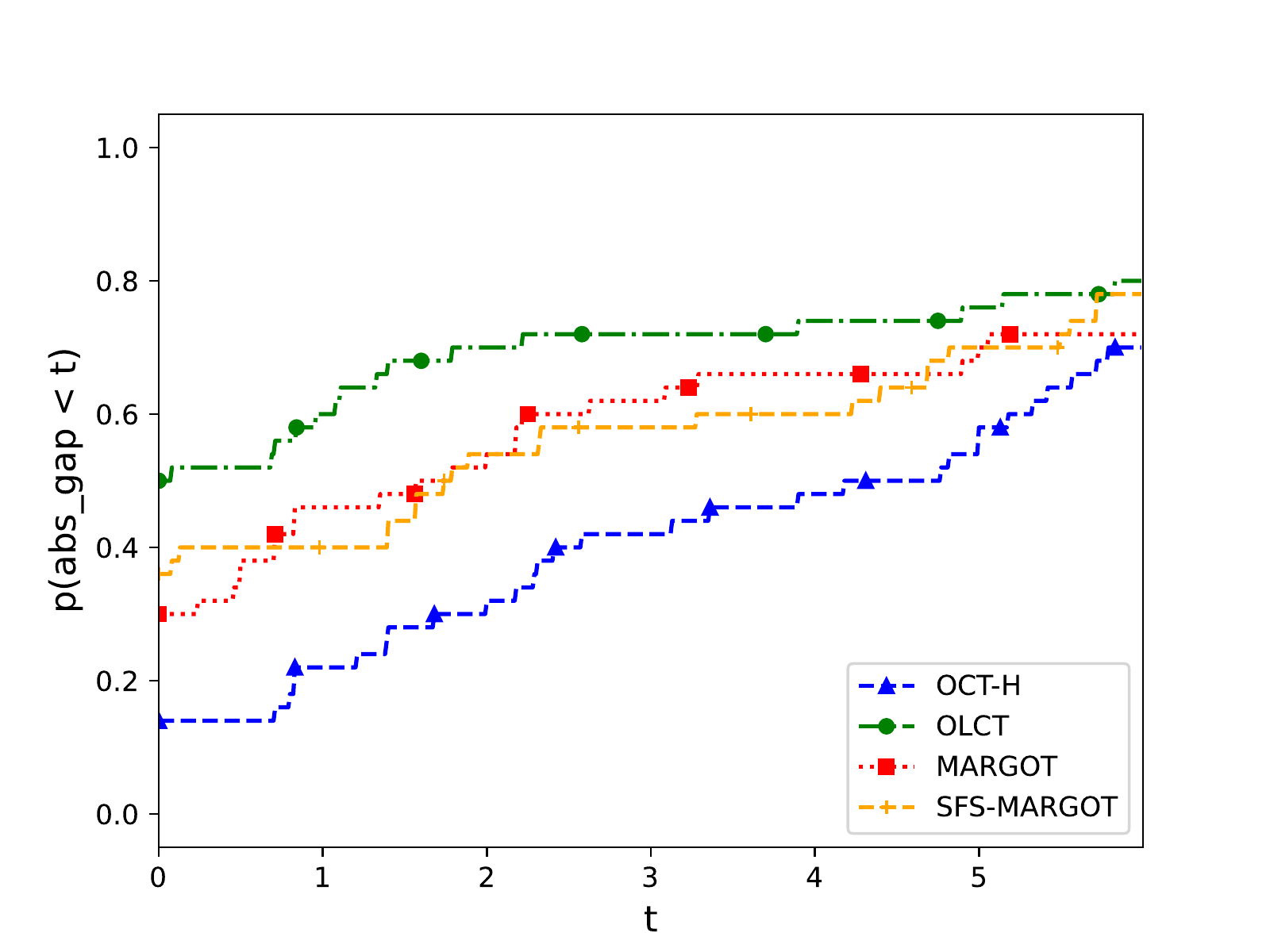}
        \caption{Cumulative distribution of the absolute gap from the best test balanced accuracy value attained by any model.}
    \label{subfig:baccuracy}
    \end{subfigure}
    \hfill
    \begin{subfigure}[t]{0.48\textwidth}
        \includegraphics[width=\textwidth]{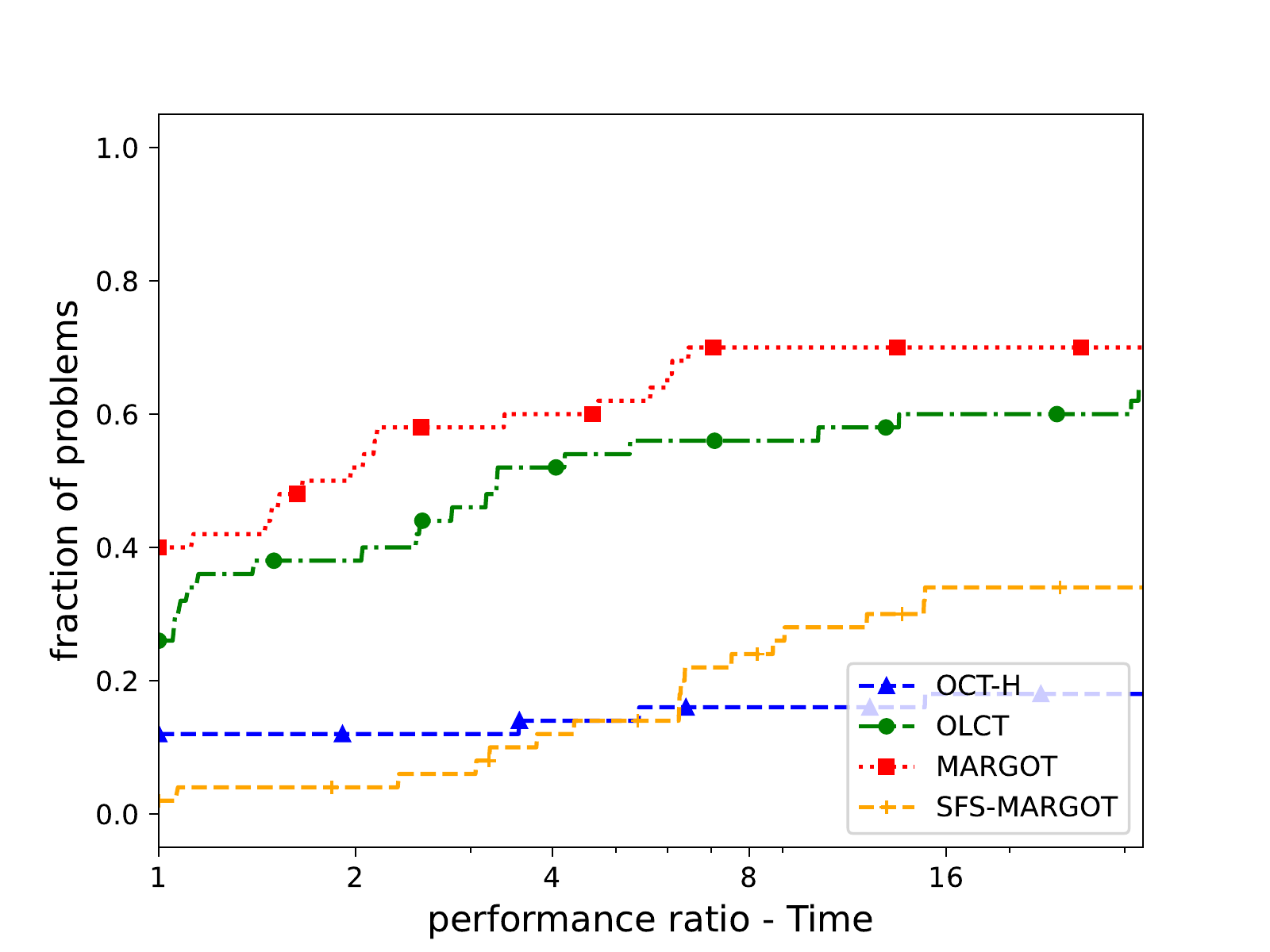}
        \caption{Performance profiles of the running times for solving the MIP models.}
    \label{subfig:time}
    \end{subfigure}
    \begin{subfigure}[t]{0.48\textwidth}
        \includegraphics[width=\textwidth]{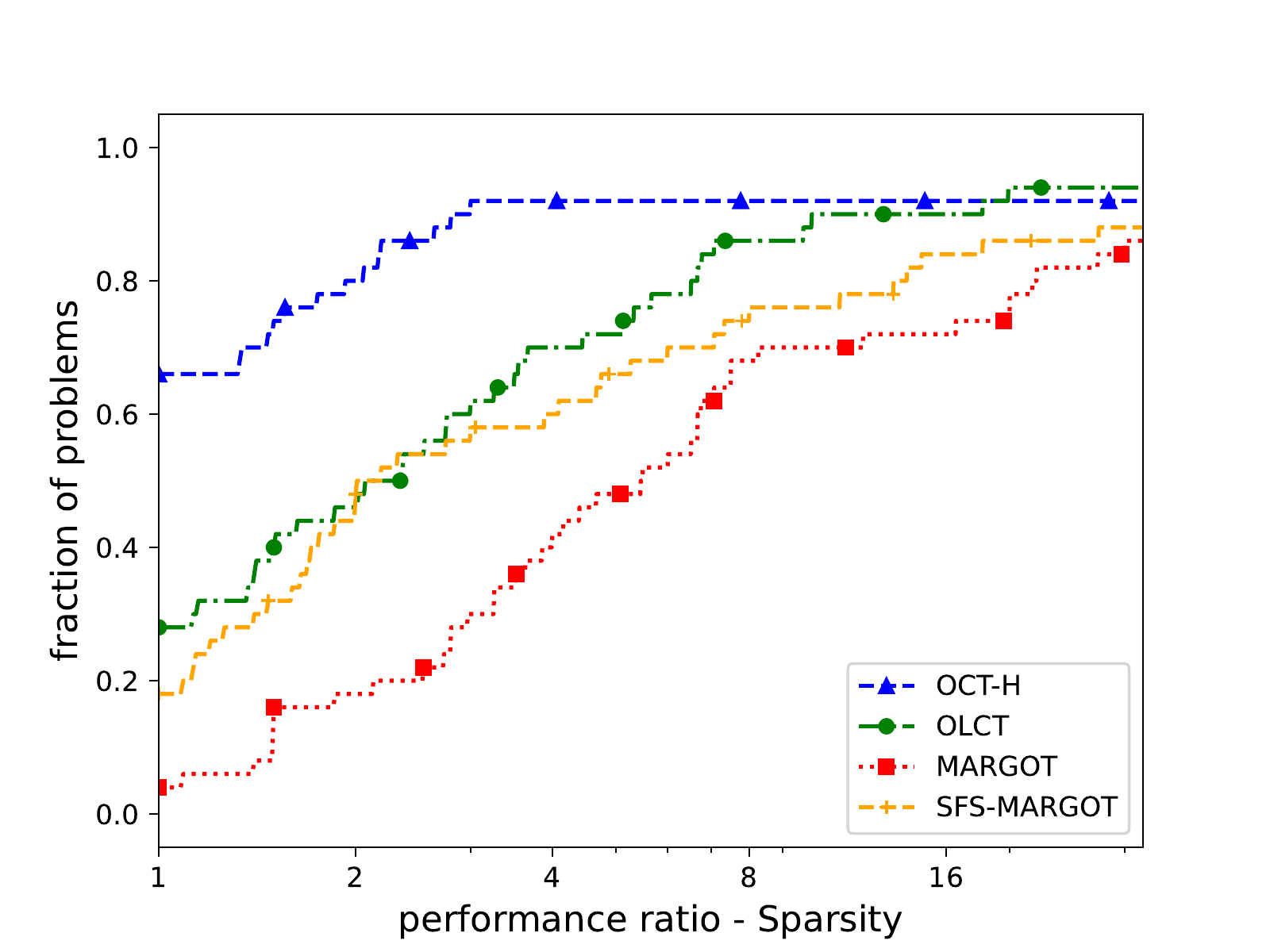}
        \caption{Performance profiles of the average number of features considered at each split of the classification tree.}
    \label{subfig:sparsity}
    \end{subfigure}
    \caption{Comparison of the performance of OLCT, MARGOT and MARGOT-SFS models, on a benchmark of 50 problem instances (10 datasets, 5 random seeds for train/test split).}
    \label{fig:results}
\end{figure}

The results of the experiment are shown in Figure \ref{fig:results} in the form of cumulative distribution of absolute gap from the best balanced accuracy on the test set and performance profiles \cite{dolan2002benchmarking} of runtime and sparsity. Sparsity is measured by the average number of features used at each node of the tree, and constitutes a proxy measure for interpretability of oblique trees.

Observing Figure \ref{subfig:baccuracy} we can infer, for each model, estimates of the probability to obtain a test balanced accuracy value distant at most $t$ points from the best one attained by any model. In other words, for any $t$, we can observe the fraction of times a model reaches an accuracy level within $t$ points from the best one. 
This kind of graph is very informative. For example, for $t=0$ we obtain the fraction of times each model is the best one among all those considered. From this point of view, we observe that our proposed model attains the top accuracy in about 50\% of cases; moreover, it also appears to be the most robust, consistently being the most likely to obtain an accuracy value close to the best one, as the gap parameter $t$ increases. Our model is thus not only most frequently the best one, but when it is not, it is still the one with the lowest probability of falling shorter than any given threshold from the best result.

Another interesting observation is that both SFS-MARGOT and MARGOT exhibit similar performances. That is, the SFS variant is able to produce sparse trees without drastically reduce the out-of-sample prediction performance. In this regard, from Figure \ref{subfig:sparsity} we can observe that OLCT outperforms both MARGOT and SFS-MARGOT in terms of sparsity, i.e., interpretability. 

The high average levels of sparsity in OLCTs branches support the effectiveness of the $\ell_1$-regularization approach to this aim. This is of course not particularly surprising, as the effects of LASSO regularization are well known. Yet, the $\ell_1$-regularization approach appears to also be able to somewhat outperform the SFS strategy based on $\ell_0$ penalization: this result was not granted and is certainly worth to be remarked, even more so taking into account that it is coupled with the efficiency advantage due to the avoided use of binary variables.

On the other hand, the OCT-H approach proves to be the best one in terms of sparsity. This result is driven by two elements within the model as described in \cite{bertsimas2017optimal}: the penalty term in the objective aimed to encourage splits considering a low number of features, and a constraint in the formulation which forces the weights of each split to have a unitary $\ell_1$ norm. However, as a consequence of the combined use of these strategies, the final model tends to be ``over-regularized'', resulting the clearly worst one in terms of balanced accuracy as shown in Figure \ref{subfig:baccuracy}.

Finally, in Figure \ref{subfig:time}, performance profiles of the running times highlight that the vanilla MARGOT is, in general, the most likely model to close the optimality gap in less than 300s. However, OLCT appears to have a comparable cost, whereas the SFS-MARGOT and the OCT-H approaches are much more computationally demanding since these latter models make use of many more binary variables. By the way, it is worth to notice that no model was able to close the optimality gap in more than 70\% of the instances with a time limit of 300$s$.

Summarizing, results highlight that our method is able to outperform other approaches in terms of balanced accuracy, to increase the interpretability, inducing sparser structures and exploiting the well known logistic properties discussed in Section \ref{subseq:interpreting}, and finally that these improvements can be achieved in competitive running times with respect to the other approaches.

\subsection{Performance Analysis with Larger Scale Problems}
The mixed-integer optimization models associated with loss-optimal classification trees grow very fast in size as the number of data points or nodes layers increases. In particular, the number of integer variables (and of constraints) grows linearly with the number of data points and exponentially with the trees depth. 

This increment is problematic from the perspective of solving the training optimization problem, as even the most efficient solvers from the state-of-the-art struggle when the number of integer variables becomes large. 

We are thus interested in conducting, at least, a preliminary assessment of the scalability of the OLCT approach compared to the behavior of MARGOT and OCT-H models. In this larger scale setting, we extended the time limit to 3600 seconds for Gurobi during the final fitting of the models, after the cross-validation phase.

First, we considered 9 additional datasets\footnote{datasets are available at \url{https://www.csie.ntu.edu.tw/~cjlin/libsvmtools/datasets/}}, whose size is reported in Table \ref{table:large}. Since the computational burden to carry out the present experiment is significant, we only considered a single train-test split for each dataset. The comparison concerns the OLCT model with the $V_0$ tangent points set, the MARGOT model and the OCT-H model, setting trees depth to 2.
The results of the experiment are reported in Table \ref{table:large}. We do not report Gurobi running times as for no instance optimality of the solution was certified. In fact, for all models, the optimality gap was consistently very close to 100\% when the time limit of 3600s was reached.

We can observe that OCT-H heavily struggled on these problems, producing meaningful models only with the least large instances of the benchmark. In the most difficult problems, neither the warm start and the global steps were able to find, within their time limits, anything better than a trivial tree forwarding all the points up to the same leaf, leading to a value of $B_\text{Acc}$ of 0.5.
On the other hand, even if unable to certify optimality, both OLCT and MARGOT were able to handle the datasets. The out-of-sample performance of the two approaches is close, with OLCT apparently having a slight advantage; yet, the size of the benchmark does not allow us to state that one model is better than the other.

\begin{table}[htbp]
\footnotesize
\centering
    \begin{tabular}{lccc}
    \hline
        & \multicolumn{3}{c}{B$_\text{Acc}$} \\ 
        \cline{2-4}
        Dataset $(n,p)$ &  OLCT - $V_0$ & MARGOT & OCT-H \\
        \hline
        a6a (11220, 122) & \textbf{74.49} & 73.57  & 50.00 \\
        german (1000, 24) &\textbf{69.76} & \textbf{69.76} & 65.60 \\
        a5a (6414, 122) & 76.72 & \textbf{77.50} &  50.00  \\
        w5a (9888, 300) & \textbf{79.91} &  78.97 & 50.00  \\
        phishing (11055, 68) & \textbf{94.44}  & 93.93  & 50.00   \\
        w4a (7366, 300) & 83.02 &  \textbf{86.62} &  50.00  \\
        splice (1000, 60) & \textbf{88.14} &  83.16 &  80.26  \\
        svmguide1 (3089, 4) & \textbf{94.98} &  94.62 &  94.52  \\
        svmguide3 (1243, 22) & 66.86 &  \textbf{67.91} &  63.62  \\\hline
    \end{tabular}
    \vspace{1em}
    \caption{Performance of depth-2 trees fit based on different OCT models on larger scale datasets (available at \url{https://www.csie.ntu.edu.tw/~cjlin/libsvmtools/datasets/}). Balanced accuracy (B$_\text{Acc}$) is reported for an 80/20 train/test split. We set 300s and 3600s as time limits for the validation and the training phase respectively.}
    \label{table:large}
\end{table}

We then proceeded to evaluate the models as the depth of the trees increased. Specifically, we conducted experiments for trees with a depth of 3.
It is important to note that, in this scenario, the cost of the experiments dramatically rises. The addition of an extra layer of nodes necessitates tuning an additional hyperparameter. Consequently, the number of hyperparameters configurations to be considered in the cross-validation phase increases accordingly, rendering the entire process time-consuming. As a result, we only considered 5 datasets from Table \ref{table:dataset} with a single train/test split.

The results of the experiment are reported in Table \ref{table:depth3}. We can observe that the loss-optimal classification tree framework remains manageable when used to train more complex tree classifiers with a depth of 3. In fact, the produced models demonstrate generalization capabilities, underlining the effectiveness of the procedure. OLCT models appear to have a slight advantage over MARGOT and OCT-H, both in terms of accuracy and ease of training. However, it is important to note that the size of the benchmark considered is too small to draw definitive conclusions in this regard.
\begin{table}[htbp]
\footnotesize
\centering
    \begin{tabular}{l|cc|cc|cc}
        &  \multicolumn{2}{c}{OLCT - $V_0$} & \multicolumn{2}{c}{MARGOT} & \multicolumn{2}{c}{OCT-H}\\ 
        \textbf{Dataset } & \textbf{B$_\text{Acc}$} & \textbf{Time} & \textbf{B$_\text{Acc}$} & \textbf{Time} & \textbf{B$_\text{Acc}$} & \textbf{Time}\\ [2pt]
        \hline
        parkinsons & \textbf{89.83} & 223.66 & 88.28 & 3600  & 83.28 & 3600\\
        wholesale & \textbf{88.69} & 3600 & \textbf{88.69} & 3600 & 87.86 & 3600\\
        haberman & 58.06 & 3600 & 54.03 & 3600 & \textbf{61.18} & 3600\\
        ionosphere & 81.78 & 3600 & \textbf{83.78} & 3600 & 80.67 & 3600\\
        spectf & \textbf{63.53} & 19.12 & 59.09 & 3600 & 53.28 & 14.9 \\
    \end{tabular}
    \vspace{1em}
    \caption{Performance of depth-3 trees fit based on different OCT models. Balanced accuracy (B$_\text{Acc}$) on test set and running times are reported on an 80/20 train/test split. We set 300s and 3600s as time limits for the validation and the training phase respectively.}
    \label{table:depth3}
\end{table}


\section{Conclusions} \label{sec:conclusion}
In this work we proposed a general framework for loss-optimal classification trees, showing how different losses can be handled through a proper definition of the slack variables.
We reviewed the recent state-of-art MIP methods for the modelling of the learning problem and we encapsulated both the misclassification loss strategy proposed in \cite{bertsimas2017optimal} and the hinge-loss approach \cite{d2022margin} in our framework.

Moreover, we provided a new formulation which employs a piece-wise linear surrogate of the log-loss for the induction of logistic multivariate classification trees. We showed how logistic splits with a standard LASSO regularization can be used to construct sparser and more interpretable trees with better generalization performances, with a reasonable computational cost. The trade-off between out-of-sample accuracy, interpretability and running time attained by our proposed approach thus appears to be optimal among the models considered in our numerical experiments.

Our work opens several directions for future contributions. Possible extensions may take into account, for example, the possibility to carry over both the general framework and the logistic approach to decision diagrams \cite{florio2022optimal} or to fit ensemble methods.
Moreover, future research shall focus on strategies to employ OLCT-like models in the multiclass setting, i.e., using multinomial regression models at each node. In the case of sparse multinomial regression, this extension was carried out for example in \cite{kamiya2019feature} with an outer approximation algorithm that exploit a dual formulation. This approach, however, is not straightforwardly applicable within a full tree structure.

\section*{Statements and Declarations}
\subsection*{Funding}
The authors declare that no funds, grants, or other support were received during the preparation of this manuscript.

\subsection*{Conflict of interest}
The authors have no competing interests to declare that are relevant to the content of this article.

\subsection*{Data availability statement}
 The datasets analyzed in the present study are available in the UCI repository \url{https://archive.ics.uci.edu/ml/index.php}
 
\subsection*{Code availability statement}
All the code developed for the experimental part of this paper is publicly available at \url{https://github.com/tom1092/Optimal-Logistic-Classification-Trees}

\bibliography{bibliography.bib}

\end{document}